\documentclass[runningheads]{llncs}
\usepackage{graphicx}
\usepackage{amsmath,amssymb}
\usepackage{color}
\usepackage{algorithm}
\usepackage{multirow}
\usepackage{algorithmic}
\usepackage{subfigure}
\usepackage{enumerate}
\usepackage[width=122mm,left=12mm,paperwidth=146mm,height=193mm,top=12mm,paperheight=217mm]{geometry}

\newcommand{\SKIP}[1]{} 

\newcommand{\mbegin} {\left [ \begin{array}}

\newcommand{\mend}   {\end{array} \right ]}

\newcommand{\detbegin} {\left | \begin{array}}
\newcommand{\detend}   {\end{array} \right |}

\newcommand{\vbegin} {\left ( \begin{array}{c}}

\newcommand{\vend} {\end{array}\right )}

\def\squareforqed{\hbox{\rlap{$\sqcap$}$\sqcup$}}

\def\qed{\ifmmode\squareforqed\else{\unskip\nobreak\hfil
	\penalty50\hskip1em\null\nobreak\hfil\squareforqed
	\parfillskip=0pt\finalhyphendemerits=0\endgraf}\fi}

\newcommand{\showeqnlabel}{
	\hbox to 0pt{\quad\quad\relax\fbox{\scriptsize\rm\eqnlblx}%
	\gdef\eqnlblx{xxxx}}} \newcommand{\eqnlblx}{}

\def\@eqnnum{\rm (\theequation)\showeqnlabel}

\newcommand{\nofig}[1]{\centerline{\bf Figure here}}

\def\mat#1{\mathchoice{\mbox{\bf$\displaystyle\tt#1$}}
	{\mbox{\bf$\textstyle\tt#1$}}
	{\mbox{\bf$\scriptstyle\tt#1$}}
	{\mbox{\bf$\scriptscriptstyle\tt#1$}}}
\def\m#1{\protect\mat #1}

\pdfoutput=1
\begin{document}
\pagestyle{headings}
\mainmatter

\title{Spatial-Temporal Union of Subspaces for Multi-body Non-rigid Structure-from-Motion} 

\titlerunning{Union of Subspaces for Multi-Body NRSFM}

\authorrunning{Suryansh Kumar, Yuchao Dai, Hongdong Li}

\author{Suryansh Kumar$^1$, Yuchao Dai$^1$, Hongdong Li$^{1, 2}$}
\institute{$^1$Research School of Engineering, The Australian National University \\ $^2$ Australian Centre for Robotic Vision. \\
	\email{ \{suryansh.kumar, yuchao.dai, hongdong.li\}@anu.edu.au}
}

\maketitle

\begin{abstract}
Non-rigid structure-from-motion (NRSfM) has so far been mostly studied for recovering 3D structure of a single non-rigid/deforming object. To handle the real world challenging multiple deforming objects scenarios, existing methods either pre-segment different objects in the scene or treat multiple non-rigid objects as a whole to obtain the 3D non-rigid reconstruction. However, these methods fail to exploit the inherent structure in the problem as the solution of segmentation and the solution of reconstruction could not benefit each other. In this paper, we propose a unified framework to jointly segment and reconstruct multiple non-rigid objects. To compactly represent complex multi-body non-rigid scenes, we propose to exploit the structure of the scenes along both temporal direction and spatial direction, thus achieving a spatio-temporal representation. Specifically, we represent the 3D non-rigid deformations as lying in a union of subspaces along the temporal direction and represent the 3D trajectories as lying in the union of subspaces along the spatial direction. This spatio-temporal representation not only provides competitive 3D reconstruction but also outputs robust segmentation of multiple non-rigid objects. The resultant optimization problem is solved efficiently using the Alternating Direction Method of Multipliers (ADMM). Extensive experimental results on both synthetic and real multi-body NRSfM datasets demonstrate the superior performance of our proposed framework compared with the state-of-the-art methods \footnote{This work was completed and submitted to ACCV on 27$^{th}$ May 2016 for review. ``The author version of the paper has been accepted by Pattern Recognition".}.
\end{abstract}

\section{Introduction}

Aiming at recovering the camera motion and non-rigid structure simultaneously from 2D images emanating from monocular cameras, non-rigid structure from motion (NRSfM) is central to many computer vision applications and has received considerable attention in recent years. This classical problem is highly under-constrained. Although existing approaches in NRSfM \cite{Bregler:CVPR-2000} \cite{Dai-Li-He:CVPR-2012} \cite{Torresani-Hertzmann:ECCV-2004} \cite{Dense-NRSFM:CVPR-2013} \cite{Akhter-Sheikh-Khan-Kanade:Trajectory-Space-2010} have presented promising results but most of these methods assume that, there is only one object undergoing non-rigid deformation in the scene. However, real world non-rigid scenes are much more complex: for example multiple persons performing different activities, soccer players in the playground, salsa dance and etc. All these real world examples constitute multi-body non-rigid deformation, which could not be explained well with the single non-rigid object assumption. Therefore, it is quite natural to extend single-body NRSfM to multi-body NRSfM where the task would be to jointly reconstruct and segment multiple 3D deforming objects over-time.

In solving the problem of multi-body NRSfM, a natural and direct two-stage process is to reconstruct non-rigid multi-body structure by applying state-of-the-art non-rigid reconstruction methods\cite{Dai-Li-He:IJCV-2013}\cite{Procrustean-Normal-Distribution:CVPR-2013} \cite{Union_Subspaces:CVPR-2014} and then segment distinct objects using subspace clustering methods such as Sparse Subspace Clustering (SSC) \cite{elhamifar2009sparse} or other clustering algorithms or vice-versa. However, by adopting such pipelines the inherent structure of the problem has never been exploited, i.e non-rigid motion segmentation provides critical information to constrain 3D reconstruction while 3D non-rigid reconstruction could also constrain the corresponding motion segmentation problem. Furthermore, since the non-rigid shape deformation actually occurs in 3D space, it is more intuitive to perform segmentation of objects in 3D space rather than on projected 2D image space.

\begin{figure}[t!]
\centering
\includegraphics[width=1.0\textwidth] {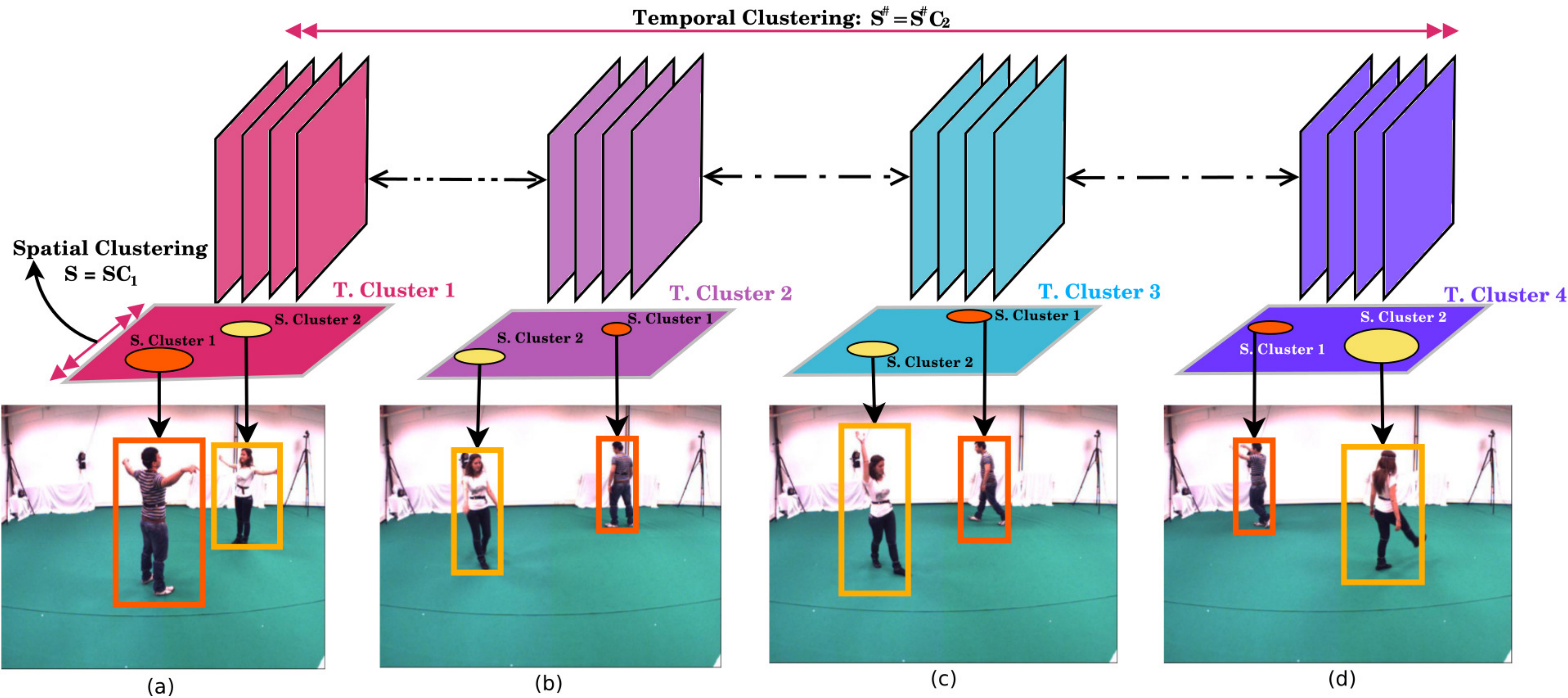}~~~
\caption{Illustration of the two clustering constraints used in our framework. We observe that, when different objects are undergoing complex non-rigid motion, the temporal clustering helps in improving the 3D reconstruction by clustering different activities over-time such as stretch, walking, jumping and etc. The spatial clustering helps in explaining the segmentation of distinct structures over images. Frames with similar activities are shown in the same colors and different subjects undergoing deformations are shown in box. Here, {\bf{T. Cluster}} refers to the Temporal cluster and {\bf{S. Cluster}} refers to the Spatial Cluster. This flow diagram demonstrates that subjects performing different activities over-time lie in distinct temporal subspace and spatial subspace, subsequently different 3D trajectories spanned by different structures lies in distinct subspace. The example images are collected from the UMPM dataset \cite{UMPM}. (Best viewed on screen in color)}
\label{fig:concept_illutration}
\end{figure}

Additionally, it is always convenient--both computationally and numerically to solve a given task using a unified approach than solving it in a sequential way. Therefore, in this paper, we propose a framework to simultaneously reconstruct and cluster multiple non-rigid shapes by exploiting the spatio-temporal correlation in data. By such approach we can explain the dynamics of non-rigid shape in a more intuitive way. Explicitly, we represent multi-body NRSfM as union of subspace both in 3D trajectory space (spatially) and 3D shape space (temporally). We use the fact that each 3D trajectory can be expressed with other trajectory only if the trajectory is from the same subspace (spatial clustering) \cite{kumar2016multi}, and each individual activity can be expressed with activity belonging to the same subspace (temporal clustering) \cite{Union_Subspaces:CVPR-2014}. A visual illustration of the spatio-temporal subspace concept is presented in Fig. \ref{fig:concept_illutration}. Concretely, spatial clustering tries to reconstruct a trajectory using affine combination of other trajectories from the same deforming object, while temporal clustering tries to explain the shape of deforming objects using affine combination of other shapes at different frame instance belonging to similar activity.

By exploiting the spatio-temporal clustering structure, our approach is able to learn the affinity matrices which naturally encode subspace information. From the affinity matrices, direct inference about number of deformable objects, different activities and membership of each sample to achieve reconstruction can be easily made. Furthermore, we exploit the fact that the connectivity between subspaces must be tight if it belongs to the same subspace and loose if belongs to different subspaces. Therefore, we propose to use a mixture of $\ell_1$ norm and $\ell_2$ norm regularization (also known as the Elastic Net \cite{Elastic_Net}), which helps in controlling the sparsity of the affinity matrices.

\paragraph{Contributions:}
\label{cn:contribution}
\begin{enumerate}
\item  We propose a joint segmentation and reconstruction framework to the challenging task of complex multi-body NRSfM by exploiting the inherent spatio-temporal union of subspace constraint.

\item  We propose to efficiently solve the resultant non-convex optimization problem based on the Alternating Direction Method of Multipliers (ADMM) method \cite{boyd2011distributed}.

\item Extensive experimental results on both synthetic and real multi-body NRSfM datasets demonstrate the superior performance of our proposed framework.
\end{enumerate}

\section{Related Works}
Multi-body structure from motion (SfM) is an important problem in computer vision. To work out this problem for rigid motion is a direct extension to elegant multi-view geometry techniques \cite{fitzgibbon2000multibody}\cite{ozden2010multibody}. However, solution to multi-body NRSfM is not straightforward, due to the difficulty in modeling complex non-rigid variations. Recent state-of-the-art in NRSfM reconstruction \cite{Dai-Li-He:IJCV-2013} has shown promising results while Zhu et al. \cite{Union_Subspaces:CVPR-2014} proposed that such an approach may fail while modeling long-term complex non-rigid motions. The work quote that Dai et al. \cite{Dai-Li-He:CVPR-2012} work is ``highly dependent on the complexity of the motion'' \cite{Union_Subspaces:CVPR-2014}. Hence, to overcome this difficulty they suggested to represent long-term non-rigid motion as union of subspace rather than a single subspace. Subsequently, Cho et al. \cite{PND-Mixture-Model:IJCV-2016} used probabilistic variations to model complex shape.

Despite the above accomplishments, NRSfM is still far behind its rigid counterpart. This gap is principally due to difficulty in modeling real world non-rigid deformation. If the deformation is irregular or arbitrary then to explain the 3D structure is nearly impossible. Nevertheless, many real world deformation can be constrained; as a result Bergler \cite{Bregler:CVPR-2000} introduced NRSfM which is considered a seminal work in NRSfM. In the work, Bergler demonstrated that non-rigid deformation can be represented by a linear combination of a set of shape basis. Following the work, several researchers tried to model NRSfM by utilizing additional constraints \cite{Torresani-Hertzmann-Bregler:PAMI-2008}, \cite{Xiao-Chai-Kanade:IJCV-2006}, \cite{Paladini:CVPR-2009}. In 2008, Akhter et al. \cite{Akhter-Sheikh-Khan-Kanade:Trajectory-Space-2010} presented a dual approach by modeling 3D trajectories. In 2009, Akhter et al. \cite{Defence-orthonormality:CVPR-2009} proved that even there is an ambiguity in shape bases or trajectory bases, non-rigid shapes can still be solved uniquely without any ambiguity. In 2012, Dai et al. \cite{Dai-Li-He:CVPR-2012} proposed a ``prior-free'' method to recover camera motion and 3D non-rigid deformation by exploiting low rank constraint only. Besides shape basis model and trajectory basis model, the shape-trajectory approach \cite{Complementary-rank-3:CVPR-2011} combines two models and formulates the problems as revealing trajectory of the shape basis coefficients. Besides linear combination model, Lee et al. \cite{Procrustean-Normal-Distribution:CVPR-2013} proposed a Procrustean Normal Distribution (PND) model, where 3D shapes are aligned and fit into a normal distribution. Simon et al. \cite{Spatiotemporal-Priors:ECCV-2014} exploited the Kronecker pattern in the shape-trajectory (spati-temporal) priors. Zhu and Lucey \cite{Convolutional-Sparse-Coding-Trajectory:PAMI-2015} applied the convolutional sparse coding technique to NRSFM using point trajectories. However, the method requires to learn an over-complete basis of 3D trajectories, prior to performing 3D reconstruction.

Recently, Russell et al. \cite{ECCV2014:video} proposed to simultaneously segment a complex dynamic scene containing a mixture of multiple objects into constituent objects and reconstruct a 3D model of the scene by formulating the problem as hierarchical graph-cut based segmentation, where the whole scene is decomposed into background and foreground objects with complex motion of non-rigid or articulated objects are modeled as a set of overlapping rigid parts.

Our method varies from the aforementioned works in the following aspects: 1) We provide a novel framework to joint segmentation and reconstruction for multiple non-rigid deformation problem; 2) We propose a simple, yet efficient and elegant optimization routine and its solution based on ADMM; 3) Our method can be applied to both sparse and dense scenarios (up to the order of ten-thousand feature tracks).

A part of this work has been published in 3DV 2016 \cite{kumar2016multi}, which addressed multi-body NRSfM by using the spatial constraint only. The work of \cite{kumar2016multi} can be viewed as a special case of the present work.

\section{Formulation}
Under our formulation, we intend to reconstruct 3D non-rigid shapes such that they satisfy both the spatio-temporal union of affine subspace constraint and the non-rigid shape constraints (low rank and spatial coherency). Let $ \m W$ $\in$ $\mathbb{R}^{2\m F \times \m P}$ represent the $measurement$ $matrix$, with $\m F$ the number of frames and $\m P$ the number of feature points. We use the $orthographic$ $camera$ model and eliminate the translation component of camera motions as suggested in \cite{Bregler:CVPR-2000}.
\begin{equation}
\label{eq:motion_shape}
\m W = \m R \m S,
\end{equation}
where $\m R = \mathrm{blkdiag}(\m R_1,\cdots,\m R_{\m F}) \in \mathbb{R}^{2 \m F\times 3 \m F}$ denotes the camera rotation matrix and $\m S$ represents the 3D shapes of deforming objects over entire frames. This classical representation for NRSfM problem \cite{Bregler:CVPR-2000} aims at recovering both the \emph{camera motion} $\m R$ and the non-rigid 3D shapes $\m S \in \mathbb{R}^{3 \m F\times \m P}$ from the 2D \emph{measurement matrix} $\m W \in \mathbb{R}^{2 \m F \times \m P}$ such that $\m W=\m R\m S$. Following the same representation to cater 2D-3D relation, we use $\|\m W - \m R \m S\|_{\m F}^2$ to infer the re-projection error.

\subsection{Representing multiple non-rigid deformations in trajectory space}
\label{ss:formulation1}
To represent multiple non-rigid objects using a single linear trajectory space does not provide compact representation of 3D trajectories \cite{Union_Subspaces:CVPR-2014}. When there are multiple non-rigid objects, each object can be characterized as lying in an affine subspace. Therefore, the 3D trajectories lie in a union of affine subspaces, which can equivalently be formulated in terms of self-expressiveness i.e,
\begin{equation}
\m S = \m S \m C_1, \mathrm{diag}(\m C_1) = \m 0, \m 1^{\m T} \m C_1 = \m 1^{\m T}.
\end{equation}
\begin{figure}
\centering
\includegraphics[width=0.6\textwidth] {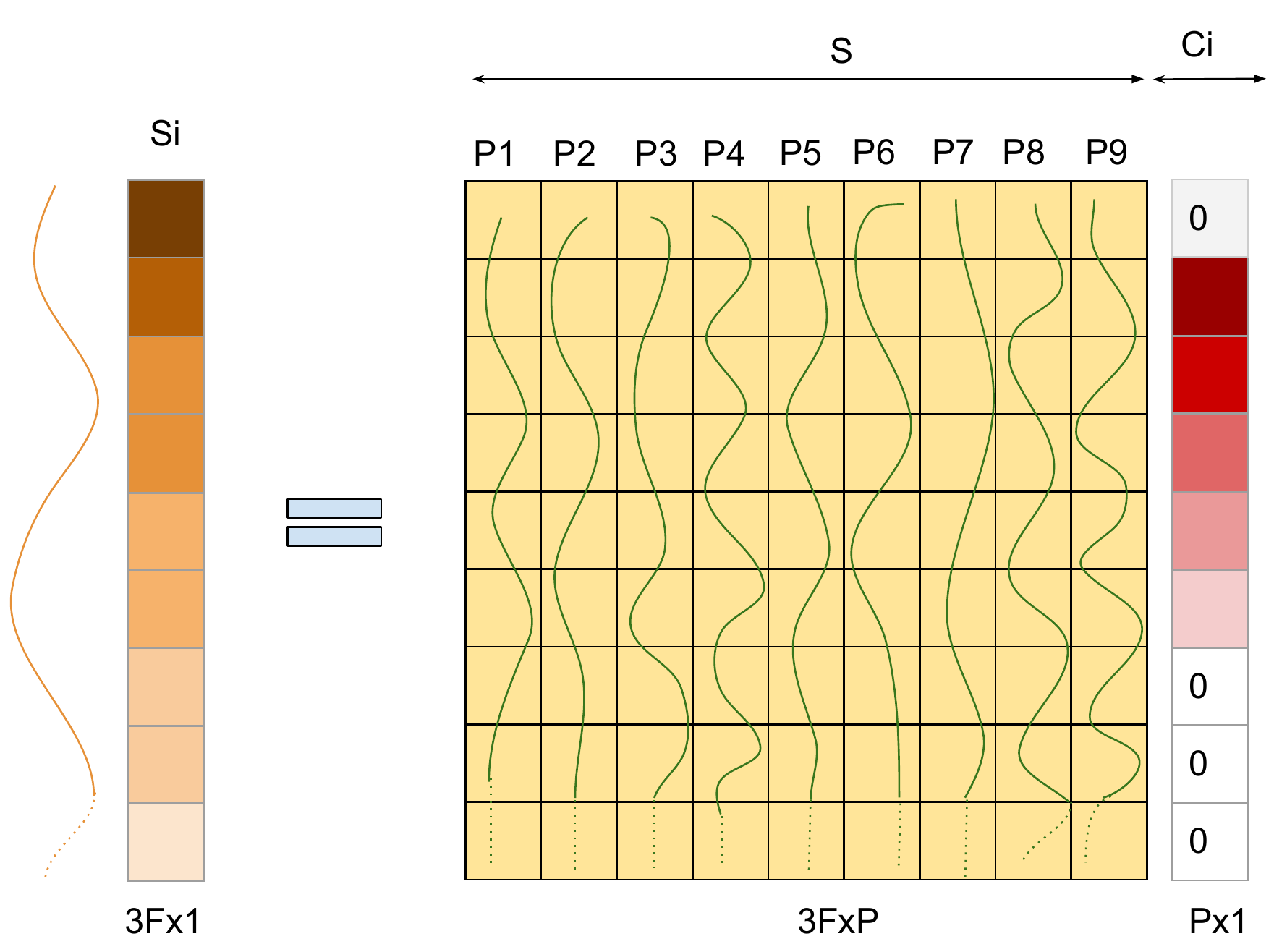}~~~
\caption{Visual illustration of the affine subspace constraint $\m S_{\m i}$ = $\m S$$\m C_{\m i}$ in trajectory space. Each column of $\m S$ is a trajectory of a 3D point (shown in green). This visualization states that a trajectory $\m S_{\m i}$ can be reconstructed using affine combination of few other trajectories. $Note :$ This pictorial representation is provided for better understanding and is only for illustration purpose. (Best viewed in color)}
\label{fig:S_Sci}
\end{figure}
where $\m S \in \mathbb{R}^{3\m F\times \m P}, \m C_{1} \in \mathbb{R}^{\m P\times \m P}$. To get rid of the trivial solution of $\m S = \m S $ or $\m C_1 = \m I$, we explicitly enforce the diagonal constraint as $\mathrm{diag} \m {(\m C_{1})} = \m 0$. As we represent each non-rigid object as lying in an affine subspace, we further enforce the affine constraint $\m 1^{\m T} \m C_{1} = 1^{\m T}$. Besides the above constraint, we also want to enforce a constraint that if the trajectories belong to the same deforming object then it must be tightly connected or loosely connected the otherwise. To cater this idea of inter-class and intra-class trajectories clustering, we use the elastic net formulation \cite{you2016oracle} to compromise between connectedness and sparsity. Combining all the constraints together, we reach the following optimization:
\begin{equation}
\begin{aligned}
& \underset{\m C_1}{\text{minimize}}~ \lambda_1\|\m C_1 \|_1 + \frac{(1-\lambda_1)}{2} \| \m C_1 \|_{\m F}^2\\
& \text{\it{subject to:}} \\
& \m S = \m S \m C_1, \mathrm{diag}(\m C_1) = \m 0, \m 1^{\m T} \m C_1 = \m 1^{\m T}, \lambda_1 \in [0, 1].
\end{aligned}
\end{equation}
A visual illustration of this idea in trajectory space for a single trajectory is provided in Fig. \ref{fig:S_Sci}. Here, $\|.\|_1$ and $\|.\|_{\m F}$ denote the $\ell_1$-norm and the Frobenius norm respectively.

\subsection{Representing multiple non-rigid deformations in shape space}

\begin{figure}
  \begin{center}
  \subfigure[\label{fig:dance_yoga_ex}]{\includegraphics[width=0.45\linewidth, height=0.4\linewidth]{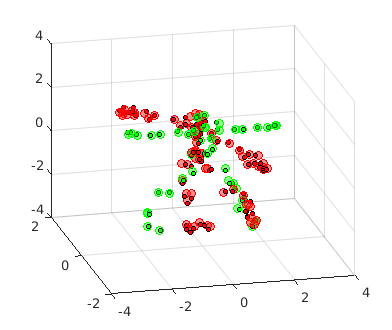}}
  \subfigure[\label{fig:shape_subspace}]{\includegraphics[width=0.50\linewidth, height=0.4\linewidth]{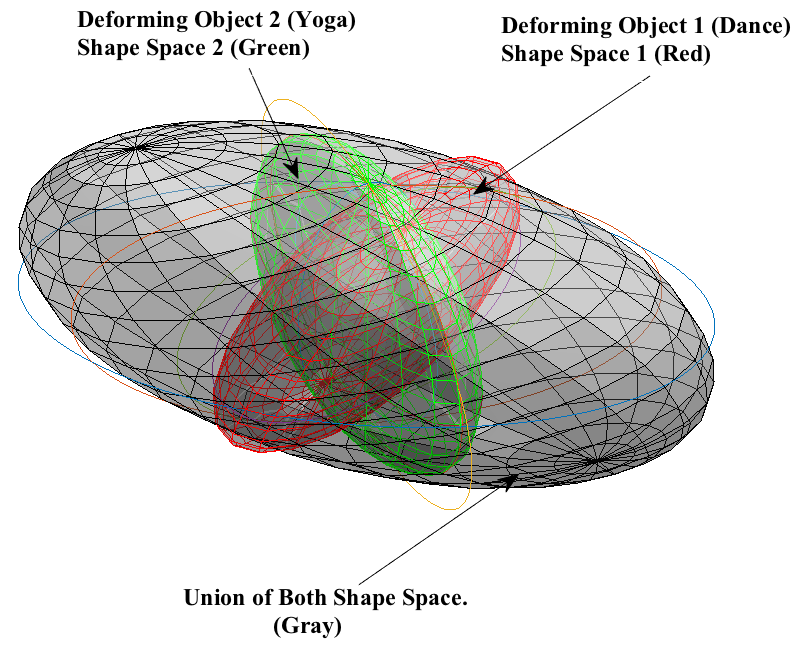}}
  \end{center}
 \caption{Visual representation of union of subspace in shape space. (a) Two different subjects are performing Dance (Red) and Yoga (Green) respectively. (b) Equivalent representation of both activities in shape space for a single frame with green ellipsoid showing the shape space for Yoga activity and red ellipsoid showing the Dance activity. It can be observed that the space spanned by different shapes performing different activities span a distinct subspace. Gray color ellipsoid shows the union of both subspaces. (Best viewed in color)}
  \label{fig:union_shape_subspace}
\end{figure}

An example complex non-rigid motion is shown in Figure \ref{fig:concept_illutration}, where the subjects are performing different activities at different time instances. Such distinct motion adheres to different local subspace and complete non-rigid motion lies in union of shape subspace. As mentioned in \cite{Union_Subspaces:CVPR-2014} such assumption leads to superior 3D reconstruction. To incorporate this concept in our formulation that different activities lie in union of affine subspaces, we express the 3D shapes in terms of self-expressiveness of frames along temporal direction.
\begin{equation}
\m S^\sharp = \m S^\sharp \m C_2, \mathrm{diag}(\m C_2) = \m 0, \m 1^{\m T} \m C_2 = \m 1^{\m T}.
\end{equation}
where $\m S^\sharp \in \mathbb{R}^{3 \m P\times \m F}$ is the reshuffled version of $\m S$ representing the per-frame 3D shape as a column vector, $\m C_{2} \in \mathbb{R}^{\m F\times \m F}.$ A visual intuition of this idea in shape space for single frame is provided in Fig. \ref{fig:union_shape_subspace}.

For temporal clustering, we also use the elastic net as regularization parameters due to similar reason mentioned in Section \ref{ss:formulation1} for $\m C_2$, thereby formulating the following optimization:
\begin{equation}
\begin{aligned}
& \underset{\m C_2}{\text{minimize}} ~\lambda_3\|\m C_2 \|_1 + \frac{(1-\lambda_3)}{2} \| \m C_2 \|_{\m F}^2\\
& \text{\it{subject to:}} \\
& \m S^\sharp = \m S^\sharp \m C_2, \mathrm{diag}(\m C_2) = \m 0, \m 1^{\m T} \m C_2 = \m 1^{\m T}, \lambda_3 \in [0, 1].
\end{aligned}
\end{equation}

\subsection{Enforcing the global shape constraint}
In seeking a compact representation for multi-body non-rigid objects, we penalize the number of independent non-rigid shapes. Similar to \cite{Dai-Li-He:CVPR-2012} and \cite{Dense-NRSFM:CVPR-2013}, we penalize the nuclear norm of the reshuffled shape matrix $\m S^{\sharp} \in \mathbb{\m R}^{3 \m P\times \m F}$, this is because the nuclear norm is known as the convex envelope of the rank function. In this way, the global shape constraint is expressed as:
\begin{equation}
\|\m S^{\sharp} \|_{*},
\end{equation}
where $\| \|_{*}$ denotes the nuclear norm of the matrix, ie, sum of singular values.

\subsection{Joint Reconstruction and Segmentation Formulation}
Putting all the above constraints (spatio-temporal union of subspace constraint and global shape constraint) together, we reach a multi-body non-rigid reconstruction and segmentation formulation:
\begin{equation}
\begin{aligned}
& \displaystyle \underset{\m S, \m C_1, \m C_2} {\text{minimize}} ~ \frac{1}{2}\| \m W - \m R \m S \|_{\m F}^2 + \lambda_1\| \m C_1 \|_1 + \frac{1-\lambda_1}{2}\|\m C_1 \|_{\m F}^2 + \lambda_2\| \m S^{\sharp}\|_{*}  + \lambda_3\| \m C_2 \|_1 + \frac{1-\lambda_3}{2}\|\m C_2 \|_{\m F}^2 \\
& \text{\it{subject to:}} \\
& \displaystyle \m S =  \m S \m C_1, \m S^{\sharp} = \m S^{\sharp}\m C_2, \\
& \displaystyle \m 1^{\m T} \m C_1 =  \m 1^{\m T}, \m 1^{\m T} \m C_2 = \m  1^{\m T}, \\
& \displaystyle \mathrm{diag}(\m C_1) = \m 0, \mathrm{diag}(\m C_2) = \m 0, \\
&\displaystyle \lambda_1, \lambda_3 \in [0, 1]. \\
\end{aligned}
\label{eq:optimization_equation_first}
\end{equation}
where $\m S^{\sharp} \in \mathbb{R}^{3 \m P\times \m F}$, $\m C_1 \in \mathbb{R}^{\m P\times \m P}$, and $\m C_2 \in \mathbb{R}^{\m F\times \m F}$. $\lambda_1, \lambda_2, \lambda_3$ are the trade-off parameters.

\section{Solution}
To solve the proposed optimization we introduce decoupling variables in Eq. \ref{eq:optimization_equation_first}, which leads to the following formulation:
\begin{equation}
\begin{aligned}
& \displaystyle \underset{\m S, \m J, \m E_1, \m E_2, \m C_1, \m C_2, \m S^{\sharp}} {\text{minimize}} ~ \frac{1}{2}\| \m W - \m R \m S \|_{\m F}^2 + \lambda_1\| \m E_1 \|_1 + \frac{1-\lambda_1}{2}\|\m E_1 \|_{\m F}^2 + \lambda_2\| \m J\|_{*}  + \lambda_3\| \m E_2 \|_1 + \frac{1-\lambda_3}{2}\|\m E_2 \|_{\m F}^2 \\
& \text{\it{subject to:}} \\
& \displaystyle \m S^{\sharp} =  \m g(\m S), \m S^{\sharp} = \m J , \\
& \displaystyle \m S =  \m S \m C_1, \m S^{\sharp} = \m S^{\sharp}\m C_2, \\
& \displaystyle \m 1^{\m T} \m C_1 =  \m 1^{\m T}, \m 1^{\m T} \m C_2 = \m  1^{\m T}, \\
& \displaystyle \mathrm{diag}(\m C_1) = \m 0, \mathrm{diag}(\m C_2) = \m 0, \\
& \displaystyle \m C_1 = \m E_1, \m C_2 = \m E_2, \\
& \displaystyle \lambda_1, \lambda_3 \in [0, 1]. \\
\end{aligned}
\label{eq:optimization_equation_first_decop}
\end{equation}
The auxiliary variables $\m E_1, \m E_2, \m J$ are introduced to simplify the derivation. $\m g(.) : \m S_{3 \m F \times \m P}$ $\rightarrow$ $\m S_{3 \m P \times \m F}^{\sharp}$ denotes the linear mapping from $\m S \in \mathbb{R}^{3 \m F \times \m P}$ to its reshuffled version $\m S^{\sharp} \in \mathbb{R}^{ \m 3\m P \times \m F}$. Specifically, S =
$\begin{bmatrix}
    \m X_{11}       & \m X_{12} & \m X_{13} & \dots & \m X_{1 \m P} \\
    \m Y_{11}       & \m Y_{12} & \m Y_{13} & \dots & \m Y_{1 \m P} \\
    \m Z_{11}       & \m Z_{12} & \m Z_{13} & \dots & \m Z_{1 \m P} \\
    \hdotsfor{5} \\
    \m X_{\m F1}       & \m X_{\m F2} & \m X_{\m F3} & \dots & \m X_{\m F \m P} \\
    \m Y_{\m F 1}       & \m Y_{\m F2} & \m Y_{\m F3} & \dots & \m Y_{\m F \m P} \\
    \m Z_{\m F 1}       & \m Z_{\m F2} & \m Z_{\m F3} & \dots & \m Z_{\m F \m P} \\
\end{bmatrix}$
and \\
$ \m S^{\sharp}$ =
$\begin{bmatrix}
    \m X_{11} \dots \m X_{1 \m P} &  \m Y_{11} \dots \m Y_{1 \m P} &  \m Z_{11} \dots \m Z_{1 \m P} \\
    \m X_{21} \dots \m X_{2 \m P} &  \m Y_{21} \dots \m Y_{2 \m P} &  \m Z_{21} \dots \m Z_{2 \m P} \\
    \dots & \dots & \dots \\
   \m X_{\m F 1} \dots \m X_{\m F \m P} &  \m Y_{\m F 1} \dots \m Y_{\m F \m P} &  \m Z_{\m F 1} \dots \m Z_{\m F \m P}
\end{bmatrix}^{\m T}$.
The first term in the above optimization is meant for penalizing re-projection error under {\it{orthographic}} projection. Under single-body NRSFM configuration, 3D shape $\m S$ can be well characterized as lying in a single low dimensional linear  subspace.   However,  when  there  are  multiple  non-rigid objects, each non-rigid object could be characterized as lying in an affine subspace. To represent this idea mathematically in shape and trajectory space respectively, we introduce $\m E_1$ and $\m E_2$.

In  addition  to  this,  to  reveal  the  intrinsic  structure  of multi-body non-rigid structure-from-motion (NRSfM), we seek  for  the  sparsest  solution both in trajectory and shape space. Consequently, we enforce the $\ell_1$ norm for $\m E_1$ and $\m E_2$. However, high sparsity may lead to misclassification of samples or trajectories. Therefore, to maintain the balance between sparsity and connectedness, we incorporate the elastic net for both $\m E_1$ and $\m E_2$. Lastly, we enforce  a  global  shape  constraint ($\|\m J\|_*$) for  compact representation of multi-body non-rigid objects by penalizing the rank of the entire non-rigid shape.

Due to the two bilinear terms $\m S = \m S \m C_1$ and $\m S^{\sharp} = \m S^{\sharp}\m C_2$, the overall optimization of Eq.-\eqref{eq:optimization_equation_first_decop} is non-convex. We solve it via the alternating direction method of multipliers (ADMM), which has a proven effectiveness for many non-convex problems and is widely used in computer vision. ADMM works by decomposing the original optimization problem into several sub-problems, where each sub-problem can be solved efficiently. To this end, we seek to decompose Eq.-\eqref{eq:optimization_equation_first_decop} into several sub-problems.

We introduce Lagrangian multipliers in the equation \eqref{eq:optimization_equation_first_decop} and reach the Augmented Lagrangian formulation for Eq.-\eqref{eq:optimization_equation_first_decop}
\begin{equation}
\begin{aligned}
& \displaystyle \mathcal{L}(\m S, \m S^{\sharp},  \m C_1, \m C_2, \m E_1, \m E_2, \m J, \{ \m Y_i\}_{\m i=1}^8) =  \frac{1}{2}\| \m W -  \m R \m S \|_{\m F}^2 + \lambda_1\| \m E_1 \|_1 + \gamma_1\|\m E_1 \|_{\m F}^2 + \lambda_2\| \m J\|_{*}  +
\\
& \displaystyle \lambda_3\| \m E_2 \|_1 + \gamma_3\|\m E_2 \|_{\m F}^2 +
<\m Y_1, \m S^{\sharp}-\m g(\m S)> + \frac{\beta}{2}\|\m S^{\sharp}-\m g(\m S)\|_{\m F}^2 +
<\m Y_2, \m S-\m S \m C_1> +
\\
& \displaystyle \frac{\beta}{2}\|\m S-\m S\m C_1\|_{\m F}^2 + <\m Y_3, \m S^{\sharp}- \m S^{\sharp}\m C_2> + \frac{\beta}{2}\|\m S^{\sharp}-\m S^{\sharp}\m C_2\|_{\m F}^2 +
<\m Y_4, \m 1^{\m T} \m C_1- \m 1^{\m T}> +
\\
& \displaystyle \frac{\beta}{2}\|\m 1^{\m T}\m C_1-\m 1^{\m T}\|_{\m F}^2 + <\m Y_5, \m 1^{\m T}\m C_2-\m 1^{\m T}> + \frac{\beta}{2}\|\m 1^{\m T}\m C_2-\m 1^{\m T}\|_{\m F}^2 + <\m Y_6, \m C_1-\m E_1> +
\\
& \displaystyle  \frac{\beta}{2}\|\m C_1-\m E_1\|_{\m F}^2 + <\m Y_7, \m C_2-\m E_2> + \frac{\beta}{2}\|\m C_2-\m E_2\|_{\m F}^2 + <\m Y_8, \m S^{\sharp}-\m J> + \frac{\beta}{2}\|\m S^{\sharp} - \m J\|_{\m F}^2,
\end{aligned}
\label{eq:Lagrangian_equation}
\end{equation}
where we define $\gamma_1 = \frac{1-\lambda_1}{2}$ and $\gamma_3 = \frac{1-\lambda_3}{2}$. $\m Y_i, i=1,\cdots, 8$ are the Lagrange multipliers. $\beta$ is the penalty parameter, where we use the same parameter for each augmented Lagrange term to simplify the derivation and parameter setting. The symbol $ <.,.> $ represents the Frobenius inner product of two matrices, i.e, the trace of the product of two matrices. For example, given two matrices $\m A, \m B \in \mathbb{R}^{\m m\times \m n}$, the Frobenius inner product is calculated as $<\m A, \m B> =$Tr$(\m A^{\m T}\m B)$.

The ADMM works by minimizing Eq.~\eqref{eq:Lagrangian_equation} with respect to one variable while fixing the others. During each iteration, we update each variable and the Lagrange multipliers in sequel. The detailed derivation for the solution is presented in the Appendix.

{\bf{Solution for S:}}
The closed form solution for $\m S$ can be derived by taking derivative of \eqref{eq:Lagrangian_equation} w.r.t to $\m S$ and equating to zero.
\begin{equation}
\begin{aligned}
\frac{1}{\beta}( \m R^{\m T} \m R + \beta \m I)\m S + \m S(\m I -  \m C_{1})(\m I -  \m C_{1}^{\m T}) = \frac{1}{\beta} \m R^{\m T} \m W + (\m g^{\m -1}( \m S^{\sharp}) + \frac{\m g^{\m -1}(\m Y_1)}{\beta } - \frac{ \m Y_2}{\beta}( \m I -  \m C_{1}^{\m T})).
\end{aligned}
\label{eq:S_complete_solution}
\end{equation}

{\bf{Solution for ${\bf{S^{\sharp}}}$:} }
The closed form solution for $S^\sharp$ can be derived by taking derivative of \eqref{eq:Lagrangian_equation} w.r.t $\m S^\sharp$ and equating to zero.
\begin{equation}
\begin{aligned}
\m S^{\sharp}(\m 2\m I + (\m I-\m C_2)(\m I-\m C_{\m 2}^{\m T}) ) = (\m g(\m S) - \frac{\m Y_{1}}{\beta}) + (\m J - \frac{\m Y_8}{\beta}) - \frac{\m Y_3}{\beta}(\m I-\m C_{2}^{\m T}).
\end{aligned}
\label{eq:S_sharp_solution}
\end{equation}

{\bf{Solution for ${\bf{\m C_{\m 1}}}$ :} } The closed form solution for $\m C_1$ can be derived as
\begin{equation}
\begin{aligned}
(\m S^{\m T}\m S + \m 1 \m 1^{\m T} + \m I)\m C_{1} = \m S^{\m T}(\m S + \frac{\m Y_{2}}{\beta}) + \m 1(\m 1^{\m T}-\frac{\m Y_4}{\beta}) + (\m E_1 - \frac{\m Y_6}{\beta}).
\end{aligned}
\end{equation}
\begin{equation}
\label{eq:Update_C1}
\m C_1 := \m C_1 - \mathrm{diag}(\m C_1),
\end{equation}

{\bf{Solution for $\m C_{2}$ :} } The closed form solution for $\m C_2$ can be derived as
\begin{equation}
\begin{aligned}
(({\m S^{\sharp}})^{\m T}\m S^{\sharp} + \m 1 \m 1^{\m T} + \m I)\m C_{2} = ({\m S^{\sharp}})^{T}\m (S^{\sharp} + \frac{\m Y_{3}}{\beta}) + \m 1(1^{\m T}-\frac{\m Y_5}{\beta}) + (\m E_2 - \frac{\m Y_7}{\beta}).
\end{aligned}
\end{equation}
\begin{equation}
\label{eq:Update_C2}
\m C_2 := \m C_2 - \mathrm{diag}(\m C_2),
\end{equation}

{\bf{Solution for $\m J$ } :} The optimization of $\m J$ given all the remaining variables can be expressed as:
\begin{equation}
\begin{aligned}
& \displaystyle \m J = \underset{\m J} {\text{argmin}} \lambda_{2}\|\m J\|_* + <\m Y_8, \m S^\sharp - \m J> + \frac{\beta}{2}\|\m S^\sharp-\m J\|_{\m F}^2.\\
& \displaystyle =  \underset{\m J} {\text{argmin}} \lambda_{2}\|\m J\|_* + \frac{\beta}{2}\|\m J - (\m S^\sharp + \frac{\m Y_8}{\beta})\|_{\m F}^2. \\
\end{aligned}
\end{equation}
A closed-form solution exists for this sub-problem. Let's define the soft-thresholding operation as $\mathcal{S}_{\tau}[x] = \mathrm{sign}(x)\max(|x|-\tau, 0)$, the optimal $\m J$ can be obtained as:
\begin{equation}
\m J = \m U\mathcal{S}_{\frac{\m \lambda_2}{\m \beta}}(\m \Sigma)\m V,
\label{eq:Jsolution}
\end{equation}
where $[\m U,$ $\m \Sigma,$  $\m V]$ = $\text{SVD}(\m S^\sharp + \frac{\m Y_8}{\m \beta})$.\\

{\bf{Solution for $ \m E_{1}$}: } The closed-form solution for $\m E_1$ can be obtained similarly:
\begin{equation}
\begin{aligned}
& \displaystyle {\m E_{1}} = \mathcal{S}_{\frac{\lambda_1}{\gamma_1 + \frac{\beta}{2}}}\left(\frac{\beta}{2\gamma_1 + \beta}(\m C_1 + \frac{\m Y_6}{\beta})\right).
\end{aligned}
\label{eq:E1solution_EN}
\end{equation}

{\bf{Solution for $\m E_{2}$ } }
The derivation for the solution of $\m E_2$ is similar to $\m E_1$.
\begin{equation}
\begin{aligned}
& \displaystyle {\m E_{2}}^* = \mathcal{S}_{\frac{\lambda_3}{\gamma_3 + \frac{\beta}{2}}}\left(\frac{\beta}{2\gamma_3 + \beta}(\m C_2 + \frac{\m Y_7}{\beta})\right).
\end{aligned}
\label{eq:E2solution_EN}
\end{equation}
Detailed derivations to each sub-problems solution are provided in \ref{ap:appendix}. Finally, the Lagrange multipliers $\{\m Y_i\}_{i=1}^8$ and $\beta$ are updated as:
\begin{equation}
\label{eq:update_Y1}
\m Y_1 = \m Y_1 + \beta(\m S^{\sharp} - g(\m S)), \m Y_2 = \m Y_2 + \beta(\m S - \m S \m C_1),
\end{equation}
\begin{equation}
\label{eq:update_Y3}
\m Y_3 = \m Y_3 + \beta(\m S^{\sharp} - \m S^{\sharp} \m C_2), \m Y_4 = \m Y_4 + \beta(\m 1^{\m T} \m C_1 - \m 1^{\m T})
\end{equation}
\begin{equation}
\label{eq:update_Y5}
\m Y_5 = \m Y_5 + \beta(\m 1^{\m T} \m C_2 - \m 1^{\m T}), \m Y_6 = \m Y_6 + \beta(\m C_1-\m E_1),
\end{equation}
\begin{equation}
\label{eq:update_Y7}
\m Y_7 = \m Y_7 + \beta(\m C_2 - \m E_2), \m Y_8 = \m Y_8 + \beta(\m S^{\sharp} - J).
\end{equation}
\begin{equation}
\label{eq:update_beta}
\beta = \min(\beta_m, \beta \rho).
\end{equation}

{\bf{Initialization:}}
Since the proposed problem is non-convex, proper initialization is required for fast convergence. In this work, we obtained rotation using \cite{Dai-Li-He:CVPR-2012} and initialized the $\m S$ matrix as pinv($\m R$)* $\m W$. $\beta_0$, $\beta_m$, $\rho$ were kept as $10^{-3}$, $10^3$, and $1.1$ respectively. The complete implementation is provided in Algorithm \ref{Algorithm 1}.

\begin{algorithm}[t!]
\caption{Multi-body non-rigid 3D reconstruction and segmentation using ADMM}
\label{Algorithm 1}
\begin{algorithmic}
\REQUIRE ~~\\
2D feature track matrix $\m W$, camera motion $\m R$, $\lambda_1$, $\lambda_2$, $\lambda_3$, $\rho>1$, $\beta_m$, $\epsilon$; \\ \vspace{0.2cm}
\hspace{-0.3cm}{\bf Initialize:} ${\m S}^{(0)}$, ${\m S^{\sharp}}^{(0)}$, $\m C_1^{(0)}$, $\m E_1^{(0)}$, $\m C_2^{(0)}$, $\m E_2^{(0)}$, $\{{\bf Y}_i^{(0)}\}_{i=1}^8 = {\bf 0}$, $\beta^{(0)}$ = $1e^{-3}$;\\ \vspace{0.2cm}
\WHILE {not converged}
\STATE 1. Update $(\m S, \m S^{\sharp}, \m E_1, \m E_2, \m C_1, \m C_2)$ by Eq.~\eqref{eq:S_complete_solution}, Eq.~\eqref{eq:S_sharp_solution}, Eq.~\eqref{eq:E1solution_EN}, Eq.~\eqref{eq:E2solution_EN}, Eq.~\eqref{eq:Update_C1} and Eq.~\eqref{eq:Update_C2}; The new value for each variable is updated over iteration, which was initialized for the first iteration.\\
\STATE 2. Update $\{{\bf Y}_i\}_{i=1}^8$ and $\beta$ by Eq.~\eqref{eq:update_Y1}-Eq.~\eqref{eq:update_beta};\\
\STATE 3. Check the convergence conditions $\| \m S^{\sharp} - \m g(\m S)\|_{\infty} \leq \epsilon$, $\| \m S - \m S\m C_1\|_{\infty} \leq \epsilon$, $\| \m S^\sharp - \m S^\sharp \m C_2\|_{\infty} \leq \epsilon$, $\| \m 1^{\m T} \m C_1 - \m 1^{\m T} \|_{\infty} \leq \epsilon$, $\| \m 1^{\m T} \m C_2 - \m 1^{\m T} \|_{\infty} \leq \epsilon$ and $\| \m C_1 - \m E_1 \|_{\infty} \leq \epsilon$, $\| \m C_2 - \m E_2 \|_{\infty} \leq \epsilon$; $\| \m S^\sharp - \m J \|_{\infty} \leq \epsilon$;  \\
\ENDWHILE
\vspace{0.2cm}
\ENSURE ~${\m C_1}$, ${\m C_2}$, ${\m E_1}$, ${\m E_2}$, $\m S, \m S^{\sharp}$.
\STATE Form an affinity matrix $\m A_1 = |\m C_1| + |\m C_1^{\m T}|$, then apply spectral clustering \cite{Spectral-Clustering:NIPS-2001} to $\m A_1$ to achieve non-rigid motion segmentation.
\end{algorithmic}
\end{algorithm}

\section{Experiments and Results}
We performed extensive experiments on benchmark data-sets that are freely available. We tested our approach on both real data and synthetic data under sparse and semi-dense scenarios. Denote $\m S^{est}$ as the estimated 3D structure and $\m S^{GT}$ as the ground-truth structure, we use the following error metrics to evaluate the performance of the approach:
\\
(i) Relative error in multi-body non-rigid 3D reconstruction
\begin{equation}
e_{3D} = \frac{1}{F}\sum_{f=1}^F\|\m S_f^{est} - \m S_f^{GT}\|_F/\| \m S_f^{GT} \|_F,
\end{equation}
(ii) Error in multi-body non-rigid motion segmentation,
\begin{equation}
e_{MS} = \frac{\mathrm{Total ~number ~of ~incorrectly ~segmented ~trajectories}}{\mathrm{Total ~number ~of ~ trajectories}}.
\end{equation}

\subsection{Experiment 1: Performance on sparse dataset}
Since our approach simultaneously reconstructs and segments multi-body non-rigid motions. Thus, we conducted the first experiment to verify the advantage of our method compared with alternative two stage approaches. To this end, we devise the following experimental setup, namely first segmenting the 2D tracks and then reconstructing each body with single body non-rigid structure-from-motion algorithm and vice-versa. Specifically, the two baseline setups are:

\begin{enumerate}[1)]
\item Baseline method 1: Single body non-rigid structure-from-motion (State-of-the-art ``block-matrix method'' \cite{Dai-Li-He:CVPR-2012} was used) followed by subspace clustering of the 3D trajectories (SSC \cite{SSC:PAMI-2013} was used), denoted as ``BMM+SSC(3D)''.
\item Baseline method 2: Subspace clustering of the 2D feature tracks (2D trajectories) followed by single body non-rigid structure-from-motion for each cluster of 2D feature tracks, denoted as ``SSC(2D)+BMM''.
\end{enumerate}

In Table ~\ref{tab:comparisons_baseline}, we provide the experimental comparisons between our method and the two baseline methods in dealing with multi-body non-rigid structure-from-motion task.

\begin{table}
\centering
  \begin{tabular}{|c|c c|c c|c c|}
    \hline
    \multirow{2}{*}{Datasets} &
      \multicolumn{2}{p{2.5cm}|}{BMM+SSC(3D)} &
      \multicolumn{2}{p{2.5cm}|}{SSC(2D)+BMM} &
      \multicolumn{2}{p{2.5cm}|}{Our Method} \\ \cline{2-7}
    & $e_{3D}$ & $e_{MS}$ & $e_{3D}$ & $e_{MS}$ & $e_{3D}$ & $e_{MS}$\\
    \hline
    Dance + Yoga & 0.045 & 0.034 & 0.058 & 0.026  & {\bf{0.045}} & 0.00\\
    Drink + Walking & 0.074 & 0.0 & 0.085 & 0.0  & {\bf{0.073}} & 0.00 \\
    Shark + Stretch & 0.024 & 0.401 & 0.098 & 0.394  & {\bf{0.021}} & 0.00\\
    Walking + Yoga & 0.070 & 0.0 & 0.090 & 0.0  & {\bf{0.066}} & 0.00\\
    Face + Pickup & 0.032 & 0.098 & \bf{0.023} & 0.098  & {{0.027}} & 0.00\\
    Face + Yoga & {\bf{0.017}} & 0.012 & 0.033 & 0.012 & 0.021 & 0.00\\
    Shark + Yoga & 0.035& 0.416 & 0.105 & 0.409 & {\bf{0.033}} & 0.00\\
    Stretch + Yoga & 0.039 & 0.0 & 0.055 & 0.0  & {\bf{0.036}} & 0.00 \\
    \hline
  \end{tabular}
  \caption{Performance comparison between our method and the two stage methods i.e first cluster and then reconstruct or vice-versa, where 3D reconstruction error ($e_{3D}$) and non-rigid motion segmentation error ($e_{MS}$) are used as error metrics. The statistics clearly shows the superior performance of our method in both 3D reconstruction and motion segmentation compared with the two stage methods.}
 \label{tab:comparisons_baseline}
\end{table}

\begin{figure*}[!htp]
  \begin{center}
  \subfigure[\label{fig:NRSFM_3D}]{\includegraphics[width=0.30\linewidth]{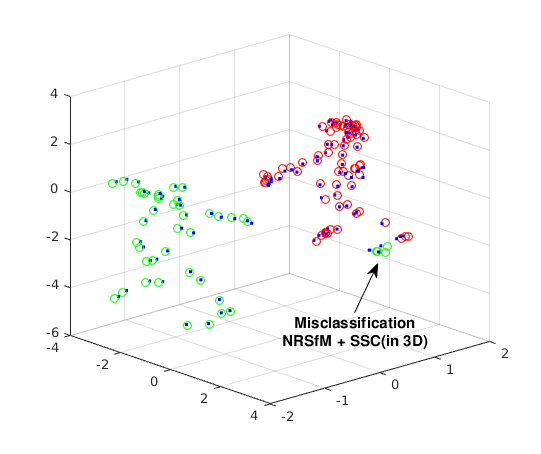}}
  \subfigure[\label{fig:SSC_NRSFM}]{\includegraphics[width=0.30\linewidth]{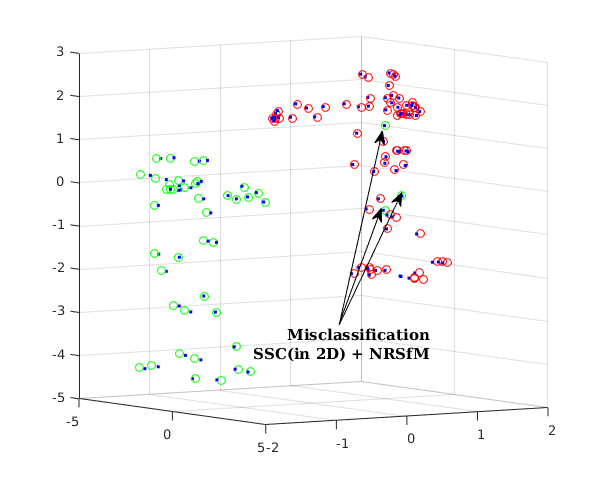}}
  \subfigure[\label{fig:our_method}]{\includegraphics[width=0.30\linewidth]{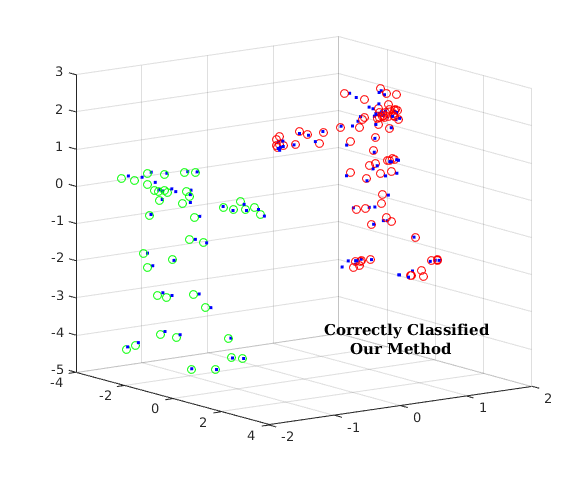}}
  \end{center}
  \caption{\small An illustration of the efficacy of our approach.  The plot shows the results on the ``Dance + Yoga'' sequence. (a) Result obtained by applying BMM method \cite{Dai-Li-He:CVPR-2012} to get 3D reconstruction and then using SSC \cite{SSC:PAMI-2013} to segment 3D points. (b)  Result obtained by applying SSC \cite{SSC:PAMI-2013} to 2D feature tracks and then using BMM \cite{Dai-Li-He:CVPR-2012} to each cluster to get 3D reconstruction. (c) Result from our simultaneous reconstruction and segmentation framework. (Best viewed on screen in color)}
  \label{fig:comparison_efficacy}
\end{figure*}

\begin{figure}[t!]
\centering
\subfigure [\label{fig:11}] {\includegraphics[width=0.3\textwidth]{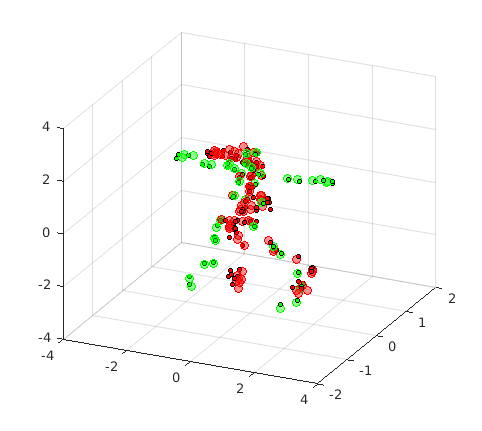}}
\subfigure [\label{fig:22}] {\includegraphics[width=0.3\textwidth]{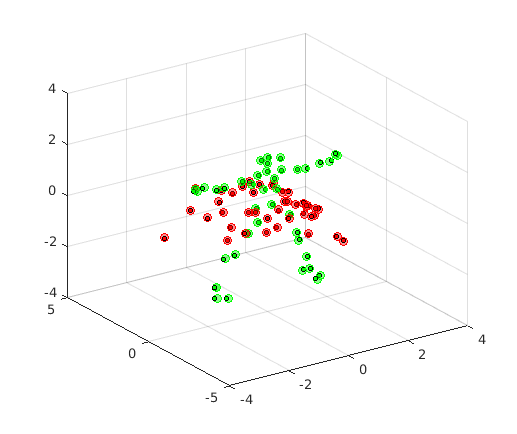}}
\subfigure [\label{fig:33}]{\includegraphics[width=0.3\textwidth]{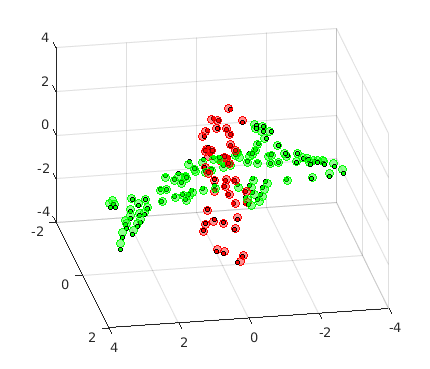}}
 \subfigure[\label{fig:44}]{\includegraphics[width=0.3\linewidth]{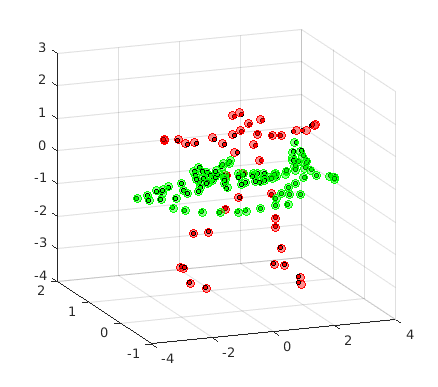}}
  \subfigure[\label{fig:55}]{\includegraphics[width=0.3\linewidth]{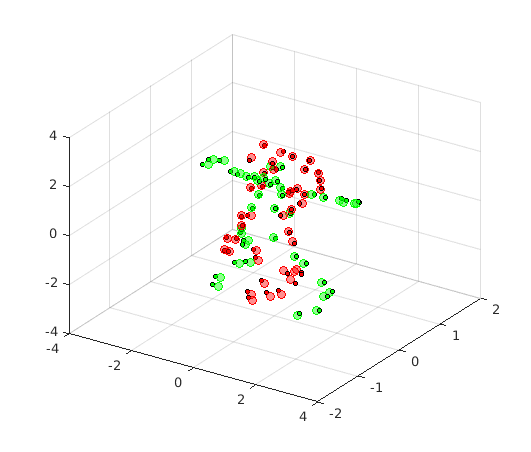}}
  \subfigure[\label{fig:66}]{\includegraphics[width=0.3\linewidth]{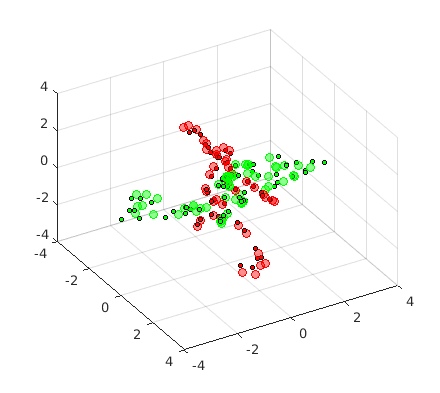}}
\caption{3D reconstruction and segmentation of different complex multi-body non-rigid motion sequences, where different objects intersect with each other. a) Dance-Yoga Sequence b) Face-Yoga Sequence c) Shark-Stretch Sequence d) Shark-Yoga Sequence e) Stretch-Yoga Sequence f) Walking-Yoga. Different colors indicate different clusters with dark small circles in the respective segments shows the ground-truth 3D points. (Best viewed in color)}
\label{fig:synthetic_sequence_linear}
\end{figure}

\begin{figure*}[!t]
\centering
\subfigure [\label{fig:1}] {\includegraphics[width=0.3\textwidth]{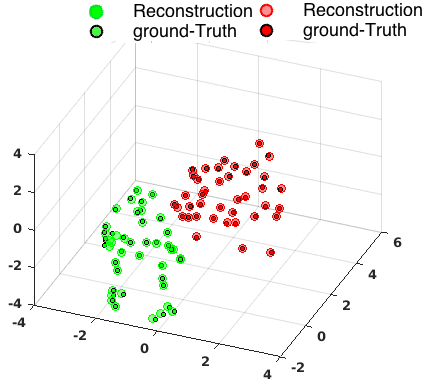}}
\subfigure [\label{fig:2}] {\includegraphics[width=0.3\textwidth]{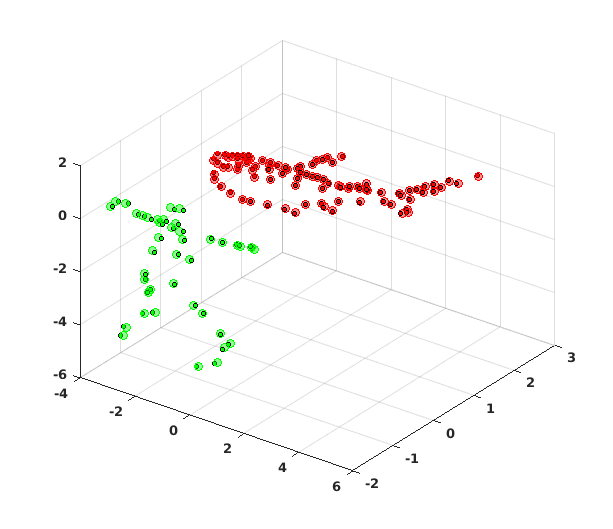}}
\subfigure [\label{fig:3}]{\includegraphics[width=0.3\textwidth]{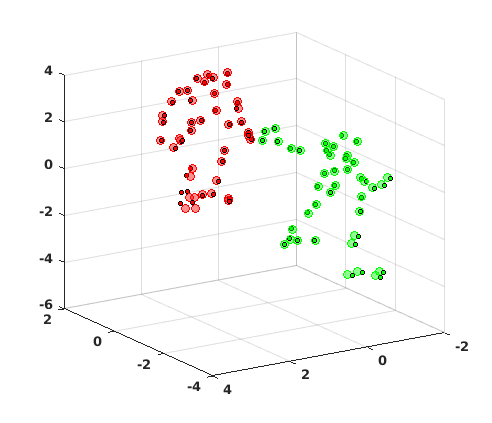}}
 \subfigure[\label{fig:dance_yoga}]{\includegraphics[width=0.3\linewidth]{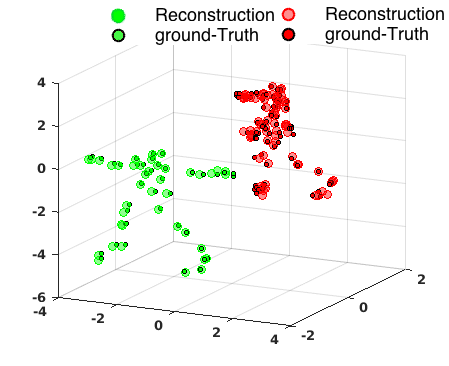}}
  \subfigure[\label{fig:p3_ball_1_result}]{\includegraphics[width=0.3\linewidth]{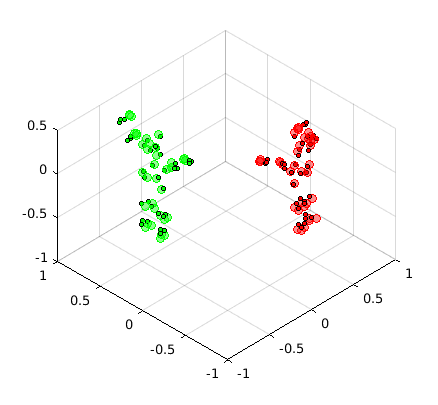}}
  \subfigure[\label{fig:p4_meet_12_result}]{\includegraphics[width=0.3\linewidth]{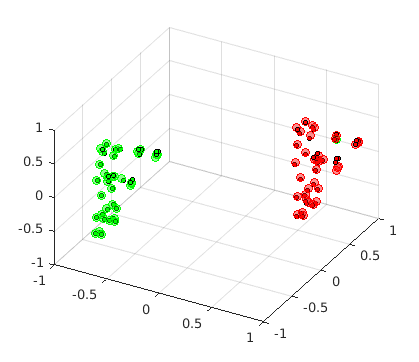}}
\caption{3D reconstruction and segmentation of different multi-body non-rigid motion sequences a) Face-Pickup Sequence b) Shark-Yoga Sequence c) Stretch-Yoga Sequence d) Dance-Yoga Sequence e) p3\_ball\_1 f) p4\_meet\_12. The non-rigid motion sequences are generated from the CMU MoCap dataset \cite{Akhter-Sheikh-Khan-Kanade:Trajectory-Space-2010}, Torresani et al. \cite{Torresani:CVPR-2001} dataset and the UMPM dataset \cite{UMPM}. Different colors indicate different clusters with dark small circles in the respective segments shows the ground-truth 3D points. (Best viewed in color)}
\label{fig:synthetic_sequence}
\end{figure*}

{\it{Comments:}} In all of these sequences, our method achieves perfect motion segmentation and better non-rigid 3D reconstruction in most of the sequences compared with the two-stage approaches--statistical value for the same sequences can be inferred from Table \ref{tab:comparisons_baseline}. Furthermore, a visual comparison is presented in Fig. \ref{fig:comparison_efficacy}, which illustrates that with the proposed framework we can procure correct features belonging to each object than the two-stage approaches.

To further test the segmentation of different deforming objects performing different activities, we simulated two synthetic experimental settings. In the first setting, we combined non-rigid objects such that they are well separated in 3D space while in the next setting the objects are intersecting with each other in 3D space. We obtained perfect segmentation results for both settings. Fig. \ref{fig:synthetic_sequence_linear} and Fig. \ref{fig:synthetic_sequence} show the qualitative segmentation and reconstruction results for the corresponding experiment. Quantitative performance comparison of segmentation with SSC \cite{elhamifar2009sparse} on synthetic sequence is presented in Table \ref{tab:comparisons_baseline} .


\subsubsection{Performance comparison of reconstruction error with state-of-the-art methods on synthetic dataset}
\begin{figure}[!htp]
\centering
\includegraphics[width=1.0\textwidth,height=0.35\textwidth] {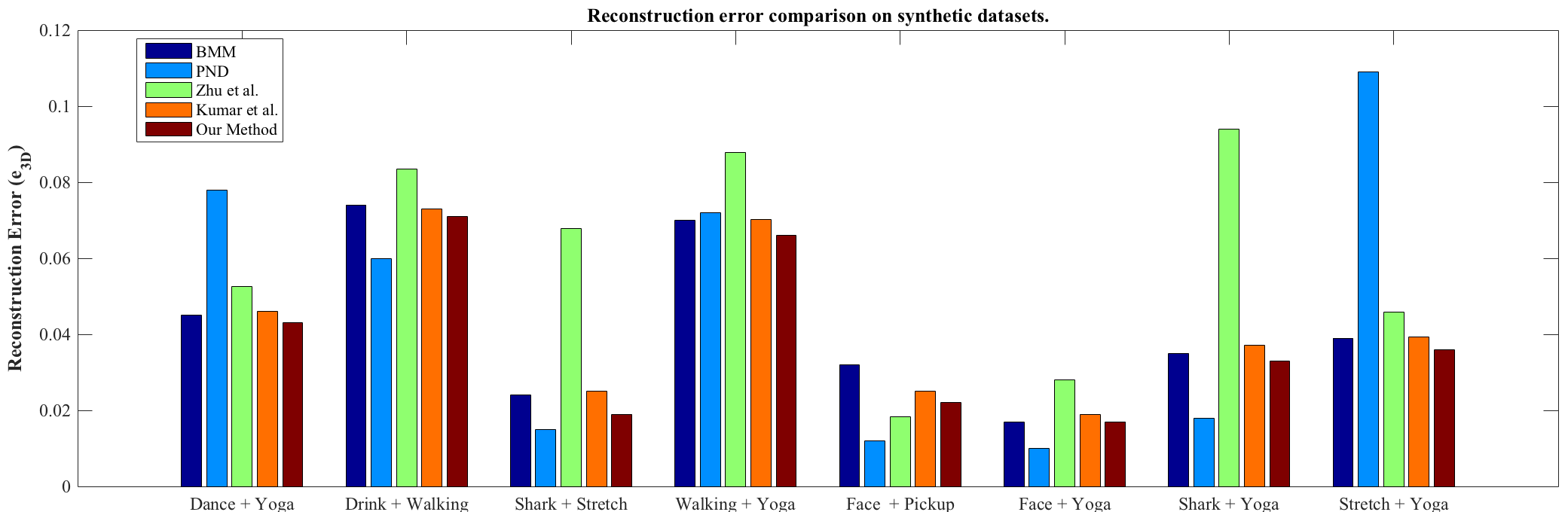}~~~
\caption{Comparison of 3D reconstruction error with other competitive methods on synthetic datasets (CMU Mocap \cite{Akhter-Sheikh-Khan-Kanade:Trajectory-Space-2010} and \cite{Torresani-Hertzmann-Bregler:PAMI-2008}). The comparison methods (BMM \cite{Dai-Li-He:CVPR-2012}, PND \cite{Procrustean-Normal-Distribution:CVPR-2013}, Zhu et al. \cite{Union_Subspaces:CVPR-2014}, Kumar et al. \cite{kumar2016multi}) present state-of-the-art approaches. Note: Code for Zhu et al. \cite{Union_Subspaces:CVPR-2014} work is not publicly available, the stats we provided here are taken from our own implementation. For exact numerical values, please refer to Table \ref{tab:Btable1} (Best viewed in color).}
\label{fig:synthetic_comparison}
\end{figure}

We compare the performance of our approach with other state-of-the-art non-rigid reconstruction methods on same data-set under similar settings. Synthetic data-set that are used for evaluating reconstruction error of multi-body non-rigid deformations are created by combining different objects from the CMU Mocap \cite{Akhter-Sheikh-Khan-Kanade:Trajectory-Space-2010} and Torresani et al. dataset \cite{Torresani:CVPR-2001}. We compare our approach with state-of-the-art non-rigid methods such as BMM \cite{Dai-Li-He:CVPR-2012}, PND \cite{Procrustean-Normal-Distribution:CVPR-2013}, Zhu et al. \cite{Union_Subspaces:CVPR-2014} and Kumar et al. \cite{kumar2016multi}. Statistical results are provided in Fig. \ref{fig:synthetic_comparison}, which clearly indicates the improvement of our method in 3D reconstruction in contrast of other approaches.

{\it{Comments:}} It can be observed from Fig. \ref{fig:synthetic_comparison} that the reconstruction error obtained by our method in comparison to other state-of-the-art is either better or close to other competing approaches on all the datasets. We would like to mention that code for Zhu et al. \cite{Union_Subspaces:CVPR-2014} is not publicly available. Therefore, we used our own implementation of this algorithm for numerical comparison. MATLAB codes for other method such as BMM \cite{Dai-Li-He:CVPR-2012} and PND \cite{Procrustean-Normal-Distribution:CVPR-2013} are freely available.

\subsection{Experiment 2:  Performance on real image dataset UMPM \cite{UMPM}.}
{\bf{UMPM :}} The Utrecht Multi-Person Motion (UMPM) dataset \cite{UMPM} is a benchmark dataset for multiple person interaction. It consists of synchronized videos with $644 \times 484$ resolution images.  Each dataset consists of long-video sequence with multiple activities and different articulated motions. Although data are provided from four view point for each category, we only used one view point for evaluation. This dataset has been used in the past as a benchmark to evaluate multi-person motion capturing technique and many state-of-the-art techniques have used it to evaluate the performance of NRSfM methods \cite{Procrustean-Normal-Distribution:CVPR-2013}, \cite{Uniqueness_Low_Rank_2012}.
\begin{figure}[!htp]
\centering
\includegraphics[width=1.0\textwidth,height=0.35\textwidth] {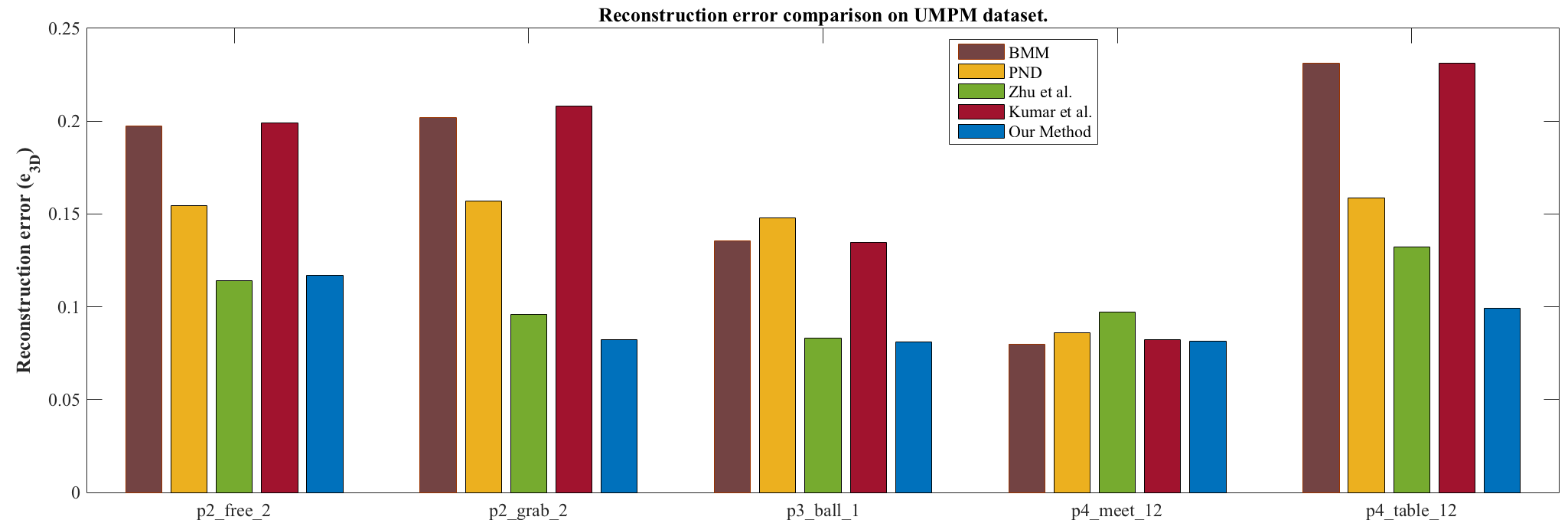}~~~
\caption{Comparison of 3D reconstruction error with other competitive methods on real image data-set(UMPM \cite{UMPM}), which is composed of complex non-rigid deformation along with different activities over-time. The comparison methods (BMM \cite{Dai-Li-He:CVPR-2012}, PND \cite{Procrustean-Normal-Distribution:CVPR-2013}, Zhu et al. \cite{Union_Subspaces:CVPR-2014}, Kumar et al. \cite{kumar2016multi}) present state-of-the-art approaches. For exact numerical values, please refer to the Table \ref{tab:Btable2} (Best viewed in color).}
\label{fig:reconstructionErrorUMPM}
\end{figure}

\subsubsection{Performance comparison of 3D reconstruction error with state-of-the-art methods on real dataset UMPM \cite{UMPM}}
Following previous works over this topic, we also used the UMPM dataset for evaluation of our method in comparison to other competing methods. We evaluated our performance on five long video sequence, which are composed of complex non-rigid motion and extensive variations of daily human actions with severe pose changes. Those sequences are  p4\_table\_12, p4\_meet\_12,  p2\_grab\_2, p2\_free\_2, and p3\_ball\_1.

\begin{figure}[!t]
  \begin{center}
  \subfigure[\label{fig:p4_table_12}]{\includegraphics[width=0.25\linewidth, height=0.2\linewidth]{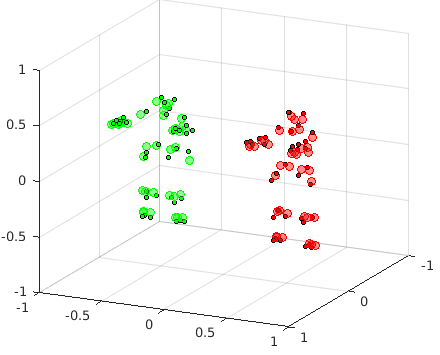}}~~
  \subfigure[\label{fig:p4_meet_12}]{\includegraphics[width=0.25\linewidth, height=0.2\linewidth]{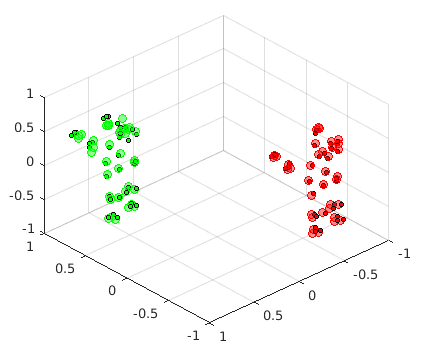}}~~
\subfigure[\label{fig:p2_grab_2}]{\includegraphics[width=0.25\linewidth, height=0.2\linewidth]{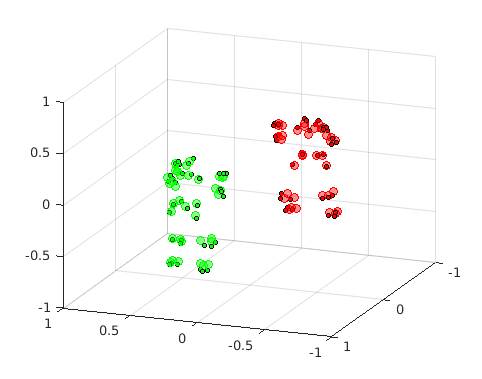}}~~
\subfigure[\label{fig:p2_free_2}]{\includegraphics[width=0.25\linewidth, height=0.2\linewidth]{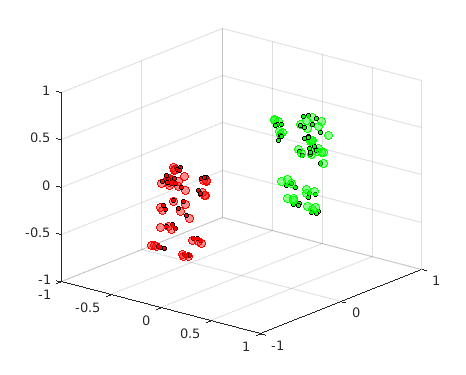}}
  \end{center}
 \caption{In (a), (b), (c), (d) larger and smaller circles shows the 3D reconstruction and ground-truth of p4\_table\_12,  p4\_meet\_12, p2\_grab\_2, p2\_free\_2 data-set respectively. Different colors show the corresponding segmentation.(Best viewed in color)}
  \label{fig:UMPM_Result1}
\end{figure}

\begin{figure}[!t]
\centering
\includegraphics[width=1.0\textwidth, height= 0.5\textheight] {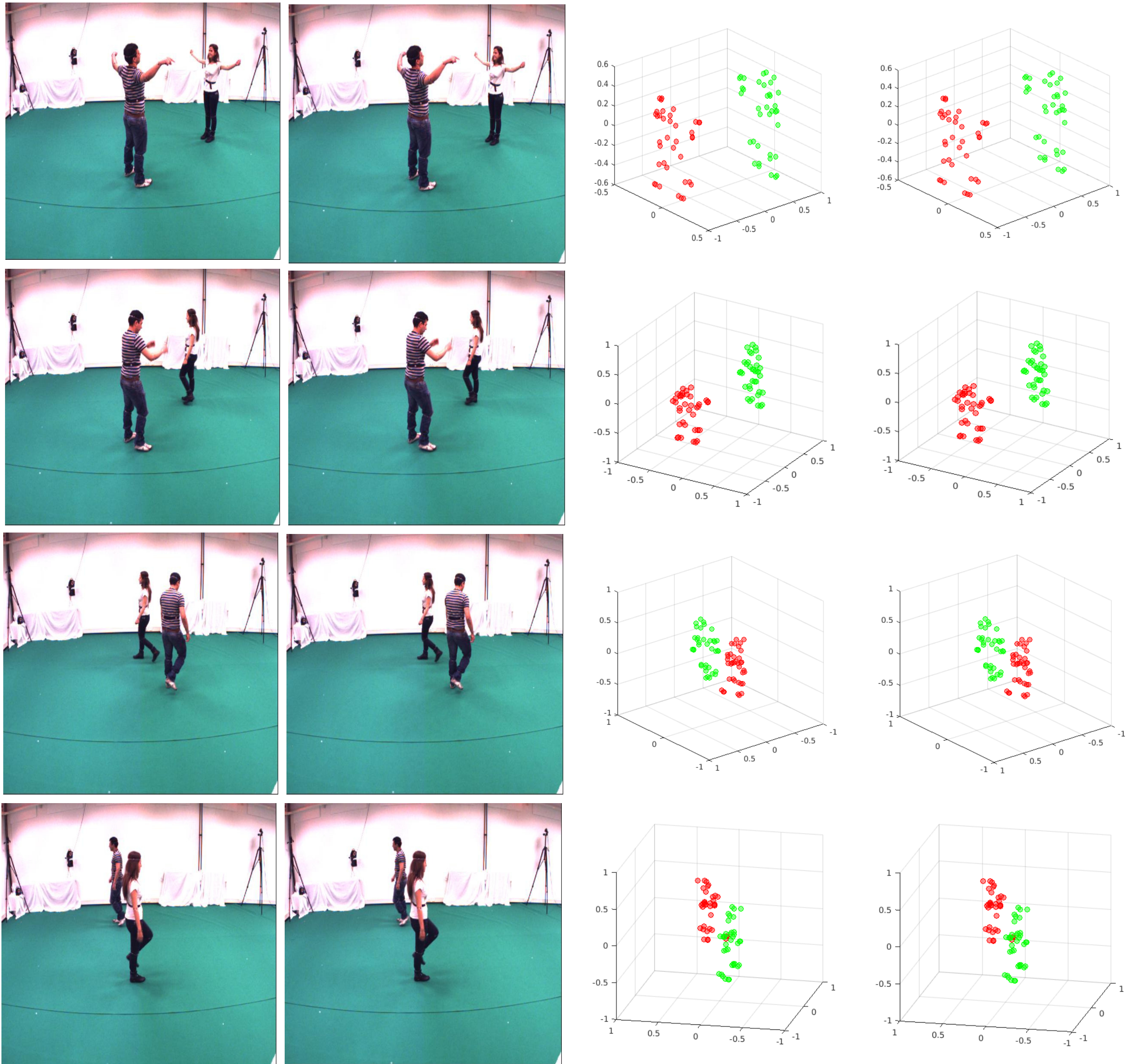}~~~
\caption{3D non-rigid reconstruction and segmentation results on p2\_free\_2 sequence of the UMPM dataset \cite{van2011umpm}. We obtained perfect segmentation and reliable 3D reconstruction over the entire video sequence which comprises of complex non-rigid deformation followed by different activities. (Best viewed in color)}
\label{fig:UMPM_Result2}
\end{figure}


{\it{Comments:}} The observations on real image experiments are very similar to the synthetic ones. In all the aforementioned data-sets, we obtained almost perfect segmentation along with reliable 3D reconstruction. Fig. \ref{fig:reconstructionErrorUMPM} demonstrates the superior 3D reconstruction performance of our method in comparison to other methods. Furthermore, qualitative results obtained using our approach on the UMPM dataset can be inferred in Fig \ref{fig:UMPM_Result1} and Fig. \ref{fig:UMPM_Result2}. Spatial and temporal affinity matrices obtained during the experiment on real sequence are analogous to synthetic sequence and therefore, similar inference can be drawn. The stats clearly indicate the superiority of our approach on 3D reconstruction, in addition it provides robust segmentation of multiple deformable objects.

\subsection{Experiment 3: Performance on dense sequences}
We also tested our method on freely available dense datasets \cite{Dense-NRSFM:CVPR-2013}. Although our method is not scalable to millions of feature tracks, for completeness of our evaluation on bench-mark dataset that consists of human facial expressions, we tested our method on the uniformly sampled version of the original sequences. We performed experiments on benchmark NRSfM synthetic and real data-set sequence \cite{Dense-NRSFM:CVPR-2013} introduced by Grag et al. This synthetic face sequence consists of four different datasets. Each sequence consists of different deformation and smooth camera rotations over time.
\begin{figure}[!htp]
\centering
\includegraphics[width=1.0\textwidth,height=0.35\textwidth] {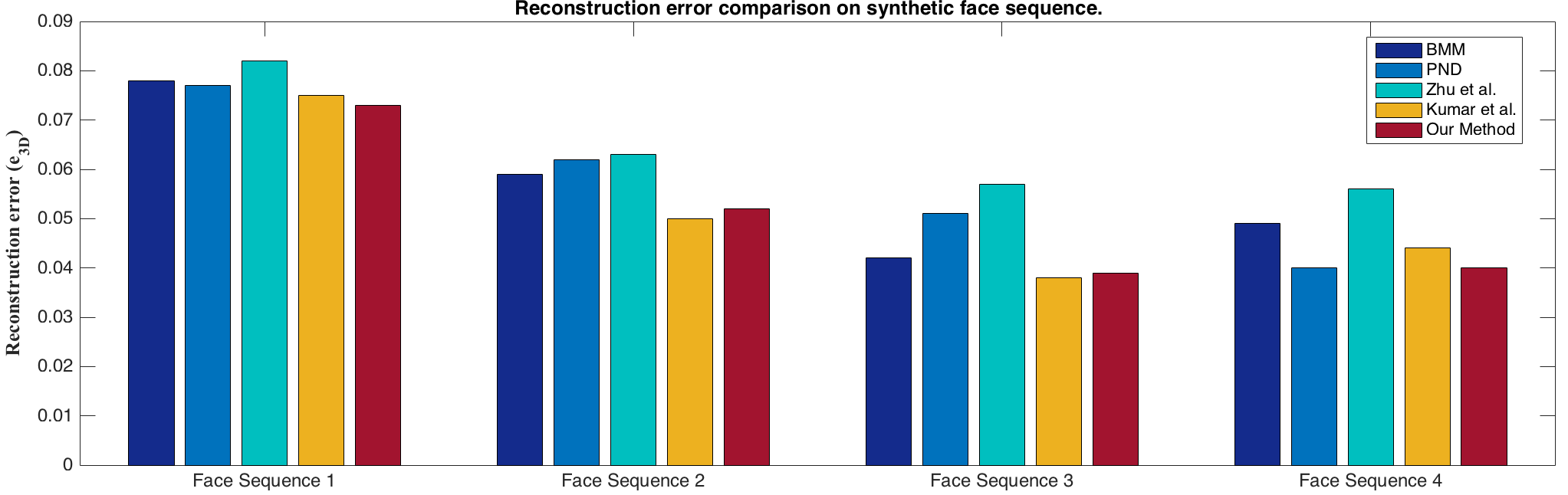}~~~
\caption{Comparison of 3D reconstruction error with other competitive methods on synthetic dense face sequence (\cite{Dense-NRSFM:CVPR-2013} ) which is composed of non-rigid face deformation of different facial expression over-time. The comparison methods (BMM \cite{Dai-Li-He:CVPR-2012}, PND \cite{Procrustean-Normal-Distribution:CVPR-2013}, Zhu et al. \cite{Union_Subspaces:CVPR-2014}, Kumar et al. \cite{kumar2016multi}) represent the state-of-the-art approaches. This comparison is made over 3275 feature tracks which is taken by uniformly sampling the dense feature tracks. For exact numerical values, please refer to the Table \ref{tab:Btable3}. (Best viewed in color).}
\label{fig:dense_3d_synthetic}
\end{figure}

\begin{figure}[t!]
\centering
\includegraphics[width= 1.0\textwidth] {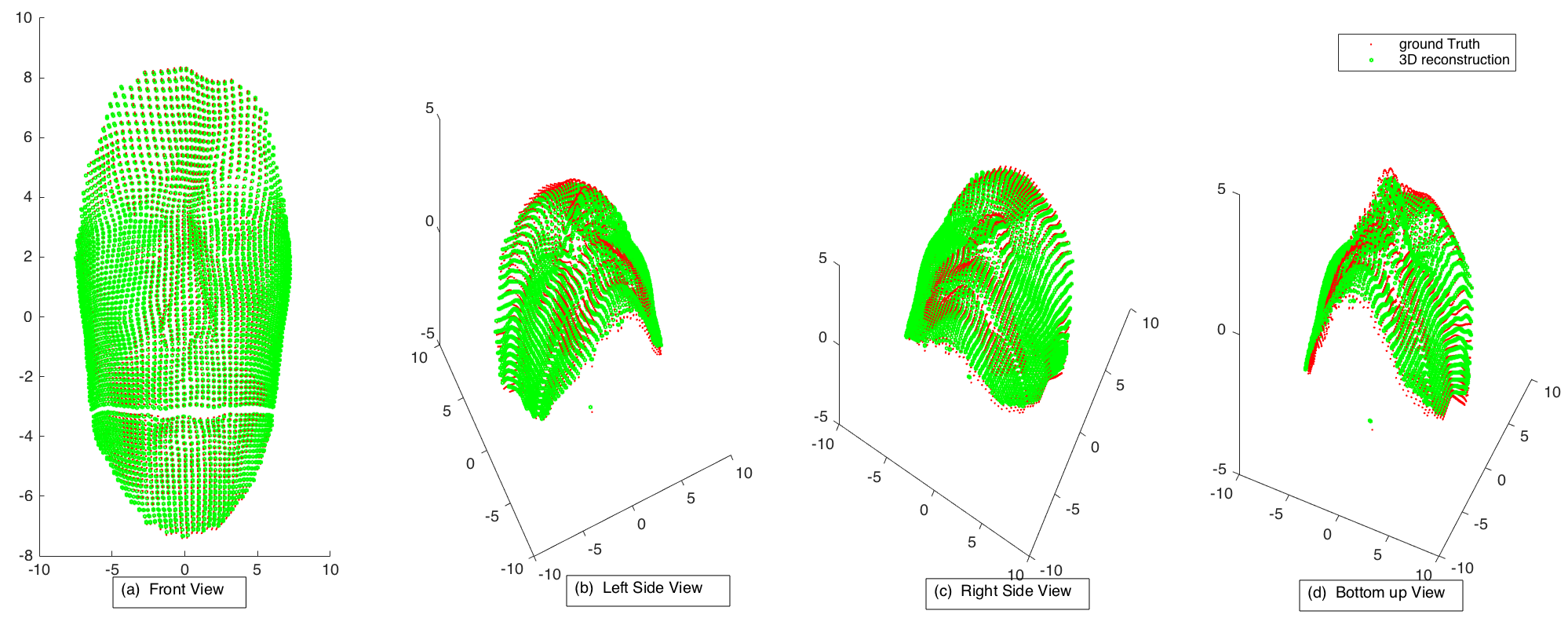}~~~
\caption{Results on synthetic face sequence \cite{Dense-NRSFM:CVPR-2013}. Red and green color show the ground-truth and reconstructed 3D structures respectively. (Best viewed in color)}
\label{fig:synthetic_face_sequence}
\end{figure}

We sampled 3275 trajectories from each synthetic face sequence to verify the performance of our approach. 3D reconstruction errors obtained over these four face sequence are shown in Fig. \ref{fig:dense_3d_synthetic}. Furthermore, Fig. \ref{fig:synthetic_face_sequence} caters the quality of reconstruction that is obtained using our method. In qualitative illustration (Fig. \ref{fig:synthetic_face_sequence}), the green dots show the reconstructed points whereas the red dots show the ground-truth 3D structure.

\begin{figure}[!htp]
\centering
  \subfigure[\label{fig:dense_seg_1}]{\includegraphics[width=0.40\linewidth]{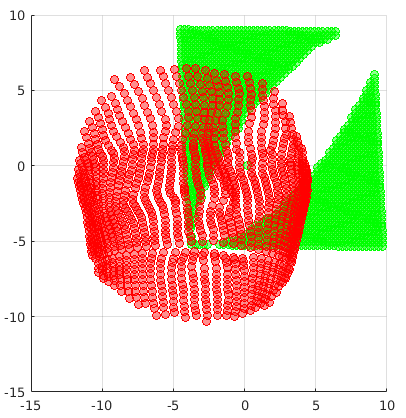}}~~
  \subfigure[\label{fig:dense_seg_2}]{\includegraphics[width=0.50\linewidth]{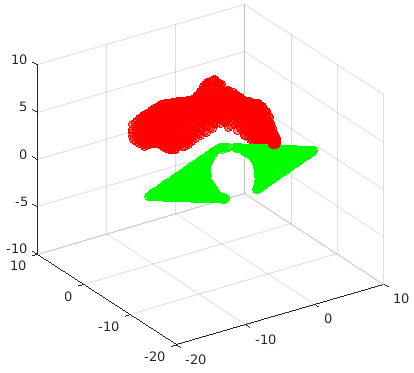}}
  \caption{(a), (b) show the front view and side view of the reconstruction and segmentation result obtained on ``Face+Background'' Sequence. This dataset was synthetically generated by combining synthetic face sequence \cite{Variational-Registration:IJCV-2013} with background as mask. (Best viewed in color)}
  \label{fig:face_back_segmentation}
\end{figure}

Face with a background is very common in real world scenarios. To test segmentation and reconstruction in such cases, we combined synthetic face with an artificial background and projected it using an orthographic camera model. We provided projected the 2D feature tracks as input to our algorithm and obtained 3D shapes as shown in Fig. \ref{fig:face_back_segmentation}. Different colors represent distinct clusters that are recovered using our method.



\subsubsection{Real face, back and heart sequence}
\begin{figure}[t!]
\centering
\subfigure[\label{fig:real_back_sequence}]
{\includegraphics[width=1.0\textwidth] {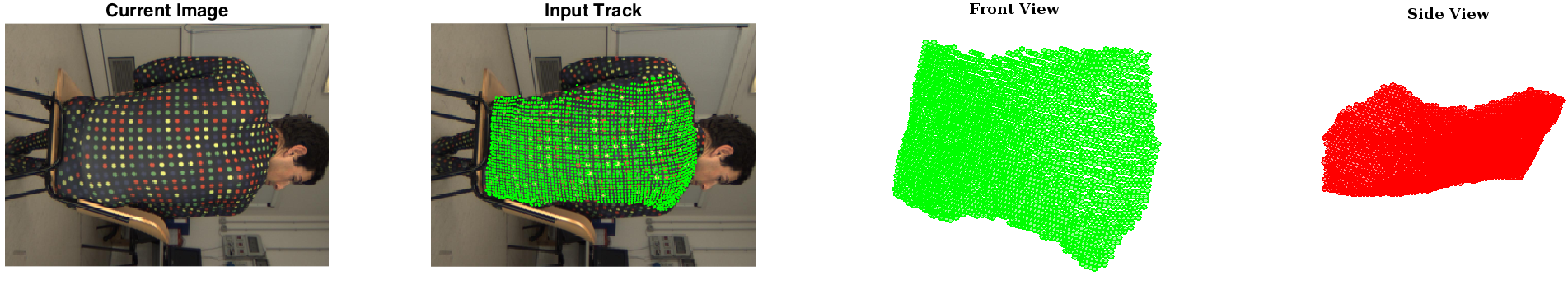}}
\subfigure[\label{fig:real_face_sequence}]
{\includegraphics[width=1.0\textwidth] {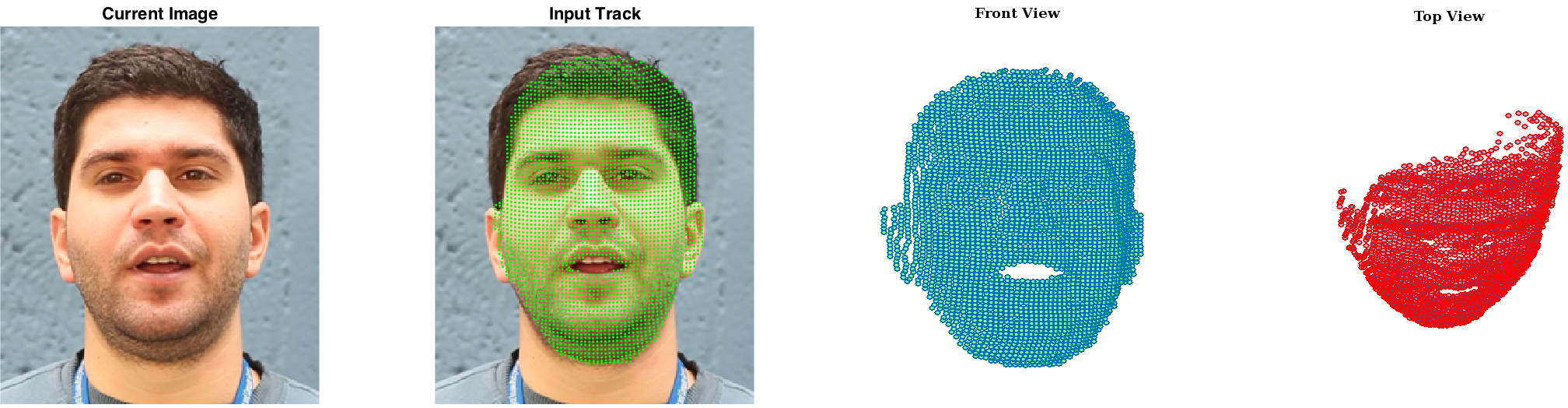}}
\subfigure[\label{fig:real_heart_sequence}]
{\includegraphics[width=1.0\textwidth, height=0.2\textheight] {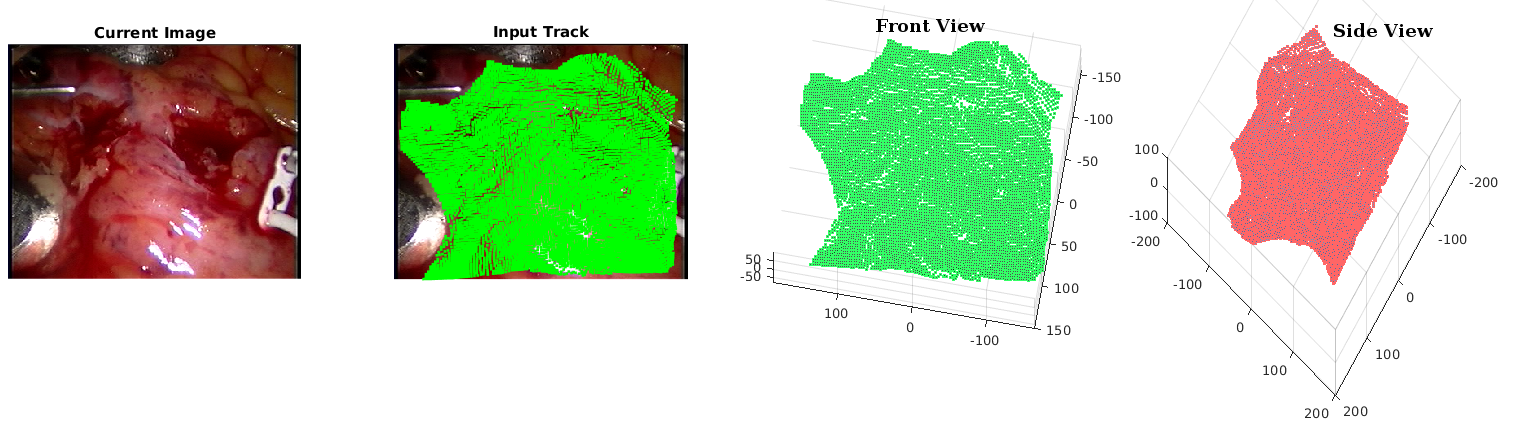}}
\caption{(a), (b), (c) shows the 3D reconstruction obtained on the Back, Face and Heart sequences respectively. Here, 2D trajectories are shown over the images to give more intuitive representation of the obtained structure. These results were obtained on uniformly sampled feature tracks. The number of feature points used for reconstruction of the Back, Face and Heart sequence are 2281, 3146 and 7546 respectively. (Best viewed in color)}
\label{fig:back_face_heart_viz}
\end{figure}

Garg et al. \cite{Dense-NRSFM:CVPR-2013} provided three monocular videos composed of face, back and heart sequence respectively. These sequences capture the natural human deformation with considerable displacements from one frame to other. In the face sequence, the subject performs day-to-day facial expression whereas in the back sequence the person is stretching and shrinking his back wearing a textured t-shirt. Lastly, this dataset also provides a challenging monocular heartbeat sequence taken during bypass surgery. Quantitative evaluation over this dataset is not provided due to the absence of 3D ground-truth. However, qualitative results obtained are shown in Fig. \ref{fig:real_back_sequence}, \ref{fig:real_face_sequence} and \ref{fig:real_heart_sequence} respectively, which demonstrates the superior performance of our method in handling these real world challenging scenarios.

\subsection{Experiment 4:  Evaluation on more than two objects.}
\begin{figure}
  \begin{center}
  \subfigure[\label{fig:result2PR}]{\includegraphics[width=0.25\linewidth, height=0.2\linewidth]{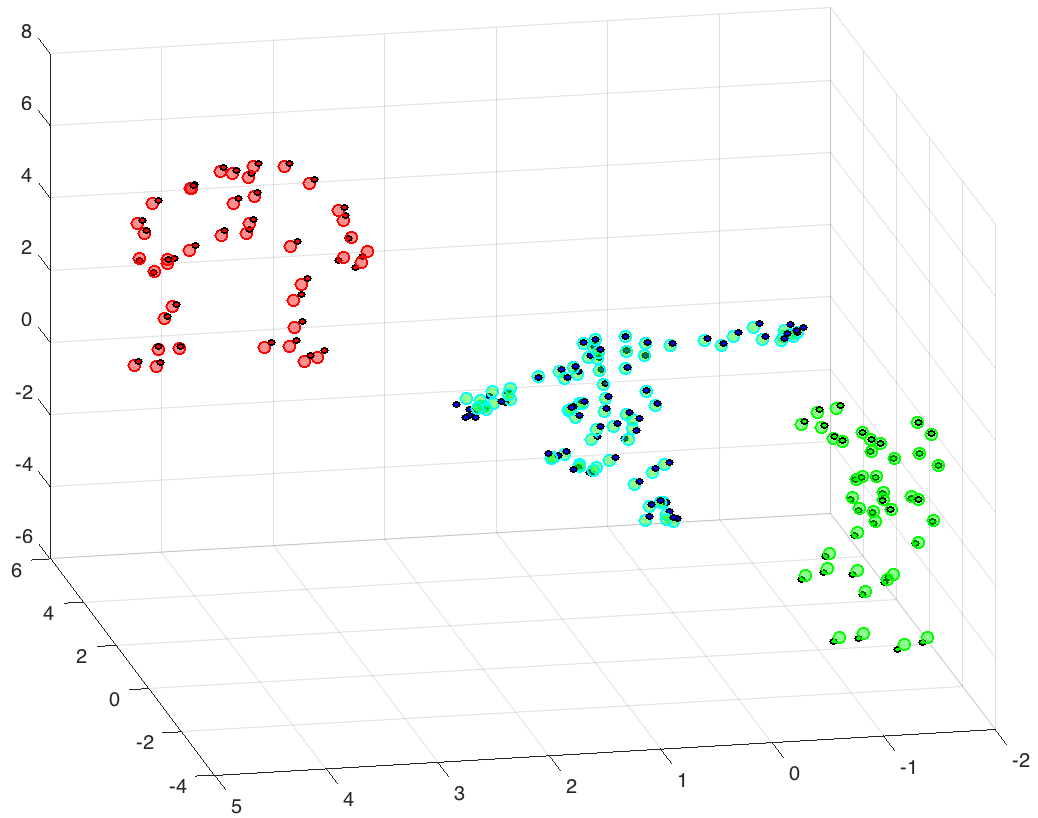}}~~
  \subfigure[\label{fig:result3PR}]{\includegraphics[width=0.25\linewidth, height=0.2\linewidth]{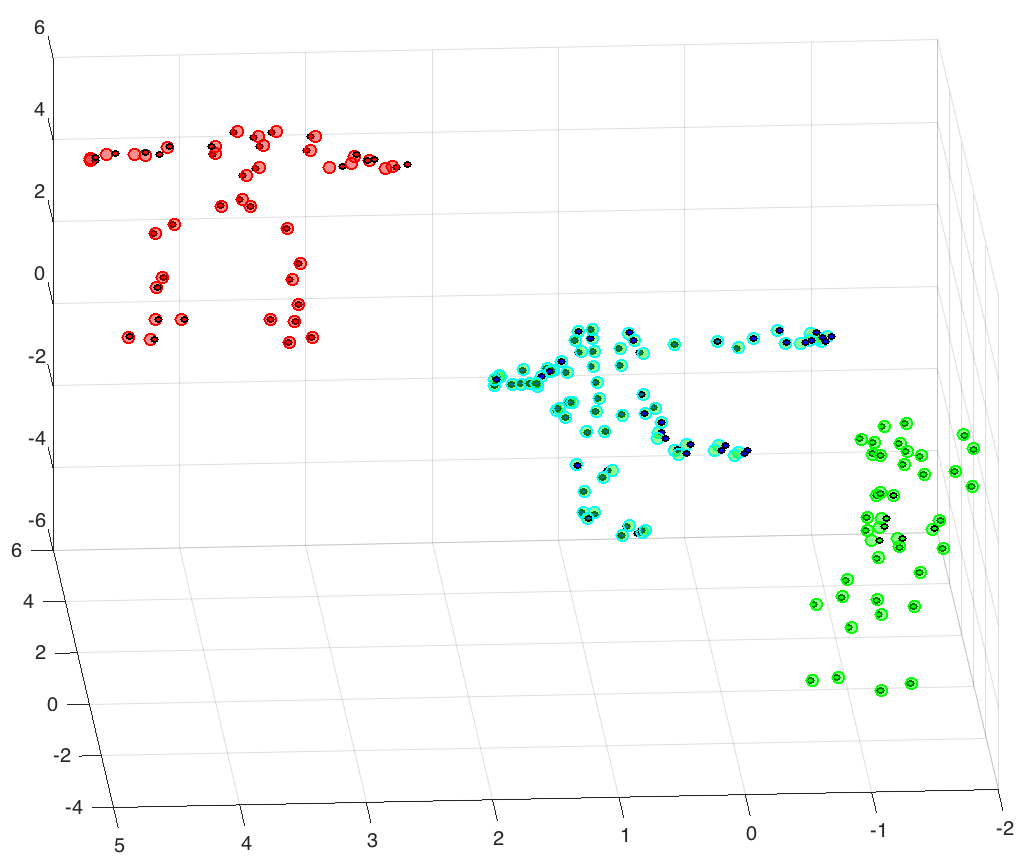}}~~
\subfigure[\label{fig:result4PR}]{\includegraphics[width=0.25\linewidth, height=0.2\linewidth]{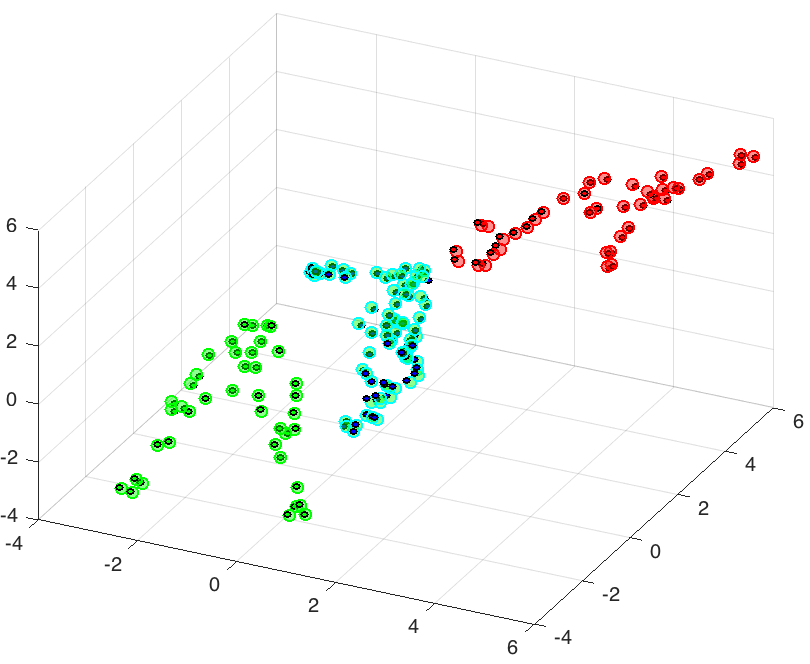}}~~
\subfigure[\label{fig:diagonalPR}]{\includegraphics[width=0.20\linewidth, height=0.18\linewidth]{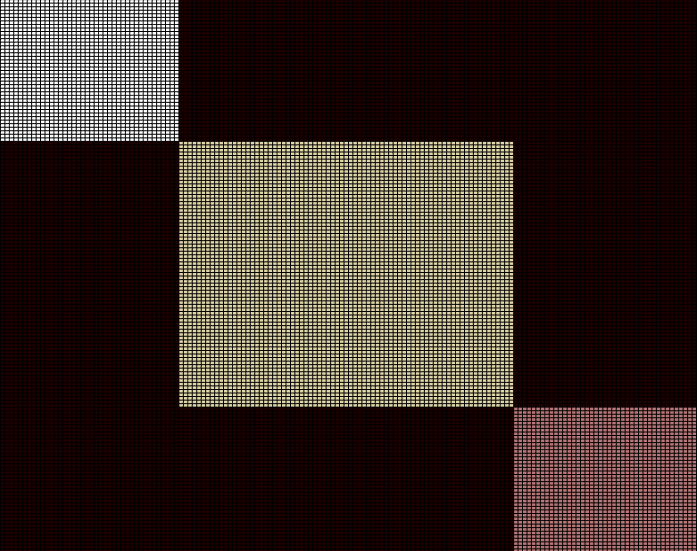}}
  \end{center}
 \caption{(a)-(c) NRSfM with segmentation results for three objects on synthetic CMU MoCap dataset \cite{Akhter-Sheikh-Khan-Kanade:Trajectory-Space-2010}. Our approach is able to reconstruct and segment each action such as stretch (red), dance (cyan) and yoga (green) faithfully with 3D reconstruction error of 0.0407. Here, different color corresponds to distinct deforming object, while dark and light color circles show ground-truth and reconstructed 3D coordinates respectively. (d) Affinity matrix obtained after spectral clustering \cite{Spectral-Clustering:NIPS-2001}. (Best viewed in color)}
  \label{fig:3obejcts}
\end{figure}

We also evaluated our method when three objects in the scene are performing complex motions over time. Adding shape clustering with trajectory clustering does not affect the segmentation, while improves reconstruction. A graphical illustration of such example and along with our obtained results in this case is shown in Fig. \ref{fig:3obejcts}

\subsection{Experiment 5: Convergence and analysis of the proposed optimization.}

\begin{figure}[t!]
\centering
\includegraphics[width=0.9\textwidth,height=0.5\textwidth] {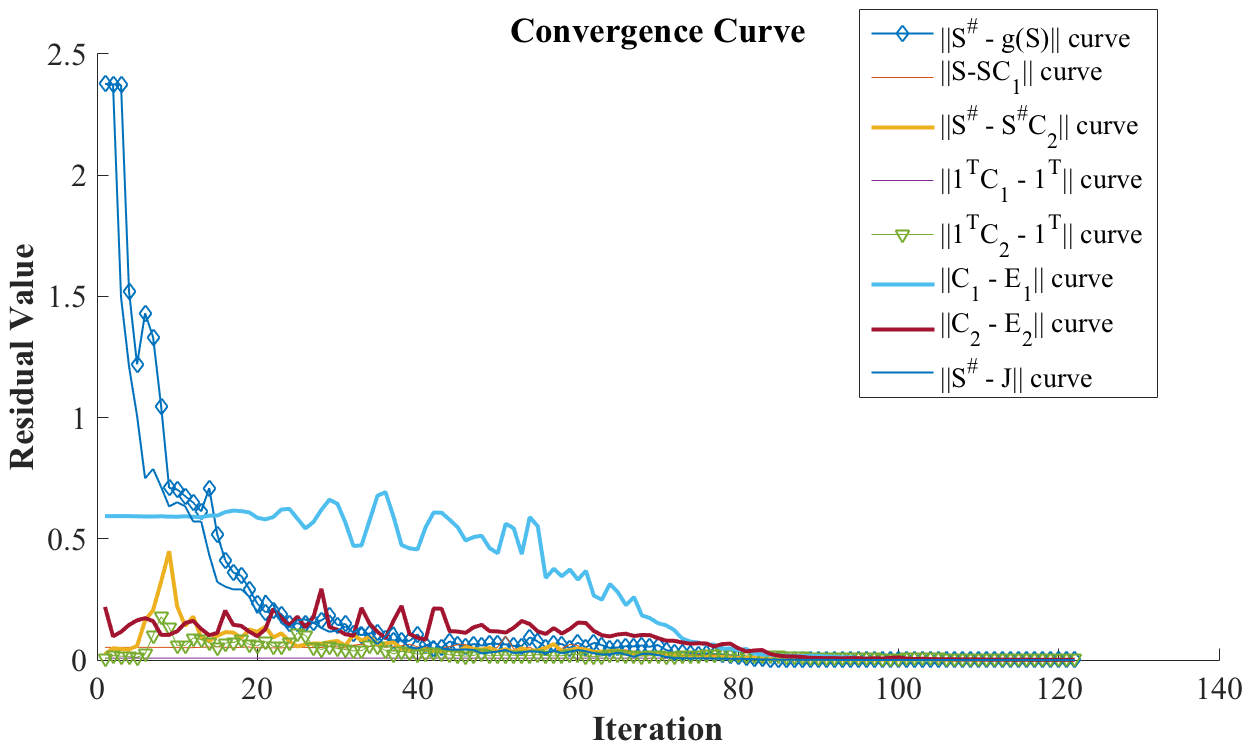}~~~
\caption{Convergence curve of the proposed optimization. Each curve represents the residual value associated with each terms shown in legends over iteration. (Best viewed in color)}
\label{fig:Convergence_Curve}
\end{figure}

Since the proposed optimization is non-convex, we conducted experiments to study the convergence and timings of our approach. {\bf{Fig. \ref{fig:Convergence_Curve}}} shows the a typical convergence curve of the proposed optimization on Shark+Yoga dataset. The optimization curve is provided only for better intuition of the algorithm. In our experiments similar convergence curves were obtained for other datasets as well. In the figure different curves shows the primal residuals for each optimization terms over iteration. The current implementation takes around 5-7 minutes for thousand feature tracks to converge on commodity desktop with MATLAB R2015b on Ubuntu 14.04 and intel core i7 processor with 16GB RAM.

\begin{figure}[!htp]
  \begin{center}
  \subfigure[\label{fig:dance_yoga_C1}]{\includegraphics[width=0.45\linewidth,  height = 0.42\linewidth]{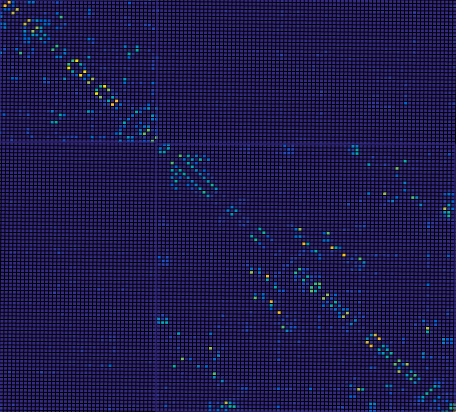}}
  \subfigure[\label{fig:dance_yoga_C2}]{\includegraphics[width=0.45\linewidth, height = 0.42\linewidth]{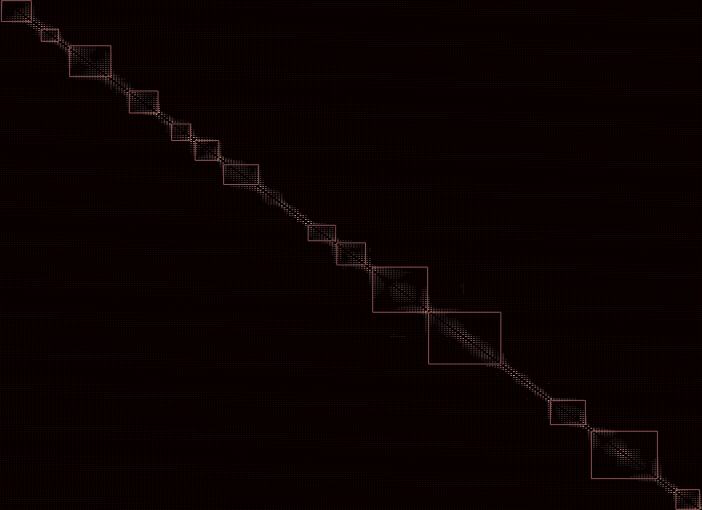}}
 \end{center}
  \label{fig:c_matrix_analysis}
  \caption{\small (a) Affinity matrix obtained on the ``Dance + Yoga'' Sequence. Clearly, it shows two block diagonal structure, corresponding to the two objects, which is an interesting observation during our experiment. Thus, number of deforming objects can be directly inferred from the affinity matrix. (b) Affinity matrix obtained with temporal clustering, it shows similar activities are encapsulated in the same block structure or captured in local subspace. (Best viewed in color)}
\end{figure}

\begin{figure}[!htp]
\centering
\includegraphics [width=0.9\textwidth,height=0.5\textwidth] {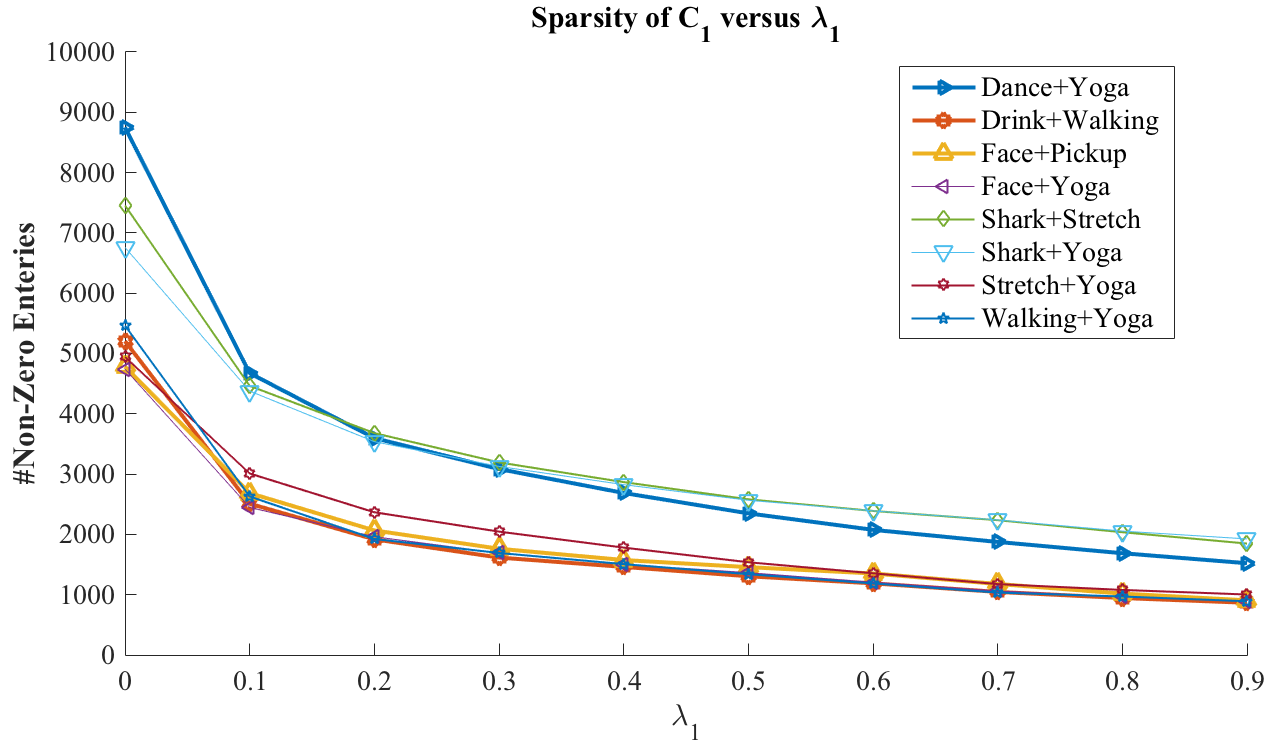}
\caption{Sparsity of $\m C_1$ matrix vs $\lambda_1$ on different sparse data-set, it can be inferred that by using a proper value of $\lambda_1$ one can control the balance between sparsity and connectedness. Similar inference can be drawn for non-zero entries of $\m C_2$ with variation in $\lambda_{3}$. (Best viewed in color)}
\label{fig:dance_yoga_C_sp}
\end{figure}

High values of $\lambda_1$ and $\lambda_3$ (say 0.6 or 0.7) during optimization may lead to higher segmentation error due to the highly sparse structure in $\m C$ matrices. The benefit of elastic net is that it provides the flexibility of trade off between the sparsity and connectedness among different classes. Mathematically it means, with elastic net we have the freedom to adjust between $\ell_1$ and $\ell_2$ minimization of the same optimization variable, which is handy in controlling the sparsity of the matrix. Figure \ref{fig:dance_yoga_C_sp} shows the sparsity of $\m C_1$ matrix with variation in $\lambda_1$ for different sparse synthetic dataset where as Fig. \ref{fig:dance_yoga_C1} and \ref{fig:dance_yoga_C2} show the affinity matrix of $\m C_1$ $\in$ $\mathbb{R}^{\m P\times \m P}$  and $\m C_2$ $\in$ $\mathbb{R}^{\m F\times \m F}$ for the Dance with the Yoga sequence. The block-diagonal structure corresponding to both deforming objects is shown in Fig. \ref{fig:dance_yoga_C1}. Clearly, the two objects span subspace that are independent of each other. In addition the obtained affinity matrix of $\m C_1$ implies that the trajectories of each individual objects are self-expressive and thus each trajectory can be represented as a linear combination of all other trajectories. Similarly, Fig. \ref{fig:dance_yoga_C2} shows similar activity spans its own subspace and therefore, frames corresponding identical action can be clustered.

\section{Conclusion}
In this paper, we proposed a novel framework to handle complex multi-body non-rigid structure from motion by exploiting spatio-temporal relation of deforming shapes, thus, providing a new way to compactly represent deformable shapes. Despite being a non-convex problem, we provided a solution to the resultant optimization using ADMM \cite{boyd2011distributed} which is effective, fast and easy to implement.
Extensive experiments on both synthetic and real benchmark datasets demonstrate that the present approach outperforms state-of-the-art non-rigid reconstruction methods, by providing competitive 3D reconstruction and highly reliable segmentation. Even though methods such as \cite{Dai-Li-He:CVPR-2012}, \cite{Paladini:CVPR-2009}, \cite{Torresani:CVPR-2001}, \cite{kumar2016multi} can handle simple variations of non-rigid deformation well, our approach provides robust reconstruction for both simple and complex multi-body deformations. In future, we plan to investigate the scalability issue with the current implementation, thus extending the framework to deal with full resolution dense reconstruction tasks (hundreds of thousands of points).


\section{Acknowledgment}
This work was supported in part by Australian Research Council (ARC) grants (DE140100180, DP120103896, LP100100588, CE140100016), Australia ARC Centre of Excellence Program on Roboitic Vision, NICTA (Data61) and Natural Science Foundation of China (61420106007).

\bibliographystyle{splncs03}
\bibliography{eccv2016submission}
\clearpage
\appendix
\section{Detailed derivation of the solution}\label{ap:appendix}
{\subsection{Solution for $\m S$}}
\begin{equation}
\begin{aligned}
& \displaystyle \m S = \underset{\m S} {\text{argmin}} \frac{1}{2}\| \m W -  \m R \m S \|_{\m F}^2 + <\m Y_1, \m S^{\sharp}-\m g(\m S)> + \frac{\beta}{2}\|\m S^{\sharp}-\m g(\m S)\|_{\m F}^2 +  <\m Y_2, \m S-\m S\m C_1>  \\ & \displaystyle + \frac{\beta}{2}\|\m S-\m S\m C_1\|_{\m F}^2.
\\
 & \displaystyle \text{We are minimizing this equation w.r.t $\m S$. Therefore, we convert the second}\\
 & \displaystyle \text{and third term in the above equation to the dimension of $\m S$.} \\
 & \displaystyle \text{$ \m S^{\sharp} = \m g(\m S) \Rightarrow \m S  = \m g^{\m -1}(\m S^{\sharp})$ (linear mapping).} \\
 & \displaystyle \text{Similarly, Lagrange multiplier $\m Y_1$ is mapped to the dimension of $\m S$.} \\
 & \displaystyle \m S = \underset{\m S} {\text{argmin}}\frac{1}{2}\| \m W -  \m R \m S \|_{\m F}^2  + \frac{\beta}{2}\|\m g^{\m -1}(\m S^{\sharp})- \m S\|_{\m F}^2 +  <\m g^{\m -1}(\m Y_1), \m g^{\m -1}(\m S^{\sharp})-\m S> \\ & \displaystyle  <\m Y_2, \m S-\m S\m C_1>  + \frac{\beta}{2}\|\m S-\m S\m C_1\|_{\m F}^2. \\
 & \displaystyle = \underset{\m S} {\text{argmin}}\frac{1}{2}\| \m W -  \m R \m S \|_{\m F}^2  + \frac{\beta}{2}(\|\m g^{\m -1}(\m S^{\sharp})\|_{\m F}^2 + \|\m S \|_{\m F}^2 - 2\m T\m r ((\m g^{\m -1}(\m S^{\sharp}))^{\m T}\m S) +\\ & \displaystyle  \m T\m r ((\m g^{\m -1}(\m Y_1))^{\m T} (\m g^{\m -1}(\m S^{\sharp})))- \m T\m r ((\m g^{\m -1}(\m Y_1))^{\m T}\m S) +  <\m Y_2, \m S-\m S\m C_1>  + \frac{\beta}{2}\|\m S-\m S\m C_1\|_{\m F}^2. \\
 & \displaystyle = \underset{\m S}{\text{argmin}}\frac{1}{2}\| \m W -  \m R \m S \|_{\m F}^2 + \frac{\beta}{2}\big(\|\m S \|_{\m F}^2 - 2\m T\m r ((\m g^{\m -1}(\m S^{\sharp}))^{\m T}\m S) -\frac{2}{\beta}\m T\m r ((\m g^{\m -1}(\m Y_1))^{\m T}\m S) \big) + \\
 & \displaystyle <\m Y_2, \m S-\m S\m C_1>  + \frac{\beta}{2}\|\m S-\m S\m C_1\|_{\m F}^2. \left\{\m S^{\sharp}, \m Y_1 \text{ are constants when minimizing over $\m S$} \right\} \\
 & \displaystyle \text{Since, adding constants to the above form will not affect the solution of S.} \\
 & \displaystyle \text{Therefore, we are adding}~~\|\m g^{\m -1}(\m S^{\sharp}) +  \frac{\m g^{\m -1}(\m Y_1)}{\beta})\|_{\m F}^2 \text{ inside the second term,} \\
 & \displaystyle \text{which will give us the form}\\
 & \displaystyle \m S = \underset{\m S} {\text{argmin}} \frac{1}{2}\| \m W -  \m R \m S \|_{\m F}^2 + \frac{\beta}{2}\|\m S - (\m g^{\m -1}(\m S^{\sharp}) +  \frac{\m g^{\m -1}(\m Y_1)}{\beta})\|_{\m F}^2 + <\m Y_2, \m S-\m S\m C_1> +
 \\
 & \displaystyle \frac{\beta}{2}\|\m S-\m S\m C_1\|_{\m F}^2.
\end{aligned}
\label{eq:simple_form_S}
\end{equation}
The closed form solution for $\m S$ can be derived by taking derivative of \eqref{eq:simple_form_S} w.r.t to $\m S$ and equating to zero.
\begin{equation}
\begin{aligned}
\frac{1}{\beta}( \m R^{\m T} \m R + \beta \m I)\m S + \m S(\m I -  \m C_{1})(\m I -  \m C_{1}^{\m T}) = \frac{1}{\beta} \m R^{\m T} \m W + \left(\m g^{\m -1}( \m S^{\sharp}) + \frac{\m g^{\m -1}(\m Y_1)}{\beta } - \frac{ \m Y_2}{\beta}( \m I -  \m C_{1}^{\m T})\right).
\end{aligned}
\end{equation}

\subsection{Solution for ${\m{S^{\sharp}}}$}
\begin{equation}
\begin{aligned}
& \displaystyle \m S^{\sharp} = \underset{\m S^\sharp} {\text{argmin}} <\m Y_1, \m S^{\sharp}-\m g(\m S)> + \frac{\beta}{2}\|\m S^{\sharp}-\m g(\m S)\|_{\m F}^2 + <\m Y_3, \m S^{\sharp}- \m S^{\sharp}\m C_2> +
\\
& \displaystyle  \frac{\beta}{2}\|\m S^{\sharp}-\m S^{\sharp}\m C_2\|_{\m F}^2 +  <\m Y_8, \m S^{\sharp}-\m J> + \frac{\beta}{2}\|\m S^{\sharp} -\m J\|_{\m F}^2. \\
& \displaystyle \text{Here, also the first two term and last two terms is condensed to a simpler}\\
& \displaystyle \text{form for mathematical convenience without affecting the final solution.}\\
& \displaystyle \m S^{\sharp} = \underset{\m S^\sharp} {\text{argmin}}~~ \m T \m r\big(\m Y_1^{\m T} \m S^{\sharp} \big) - \m T \m r\big(\m Y_1^{\m T} \m g(\m S) \big) + \frac{\beta}{2}\big(\|\m S^{\sharp}\|_{\m F}^2 + \|\m g(\m S)\|_{\m F}^2 -2 \m T \m r ((\m S^{\sharp})^{\m T} \m g (\m S)) \big)\\
& \displaystyle  + <\m Y_3, \m S^{\sharp}- \m S^{\sharp}\m C_2> + \frac{\beta}{2}\|\m S^{\sharp}-\m S^{\sharp}\m C_2\|_{\m F}^2 + \m T \m r\big(\m Y_8^{\m T} \m S^{\sharp} \big) - \m T \m r\big(\m Y_8^{\m T} \m J \big) + \frac{\beta}{2}\big(\|\m S^{\sharp}|_{\m F}^2 + \|\m J\|_{\m F}^2 + \\
& \displaystyle -2 \m T \m r \big( (\m S^{\sharp})^{\m T} \m J \big). \\
& \displaystyle \text{Since, we are minimizing over $\m S^{\sharp}$. The terms which are not dependent on $\m S^{\sharp}$}\\
& \displaystyle \text{can be considered as constants, which gives us: }\\
& \displaystyle \m S^{\sharp} = \underset{\m S^\sharp} {\text{argmin}} \frac{\beta}{2}\big(\|\m S^{\sharp}\|_{\m F}^2 - 2 \m T \m r (\m S^{\sharp})^{\m T}( \m g(\m S) - \frac{\m Y_1}{\beta}) \big) + <\m Y_3, \m S^{\sharp}- \m S^{\sharp}\m C_2> + \frac{\beta}{2}\|\m S^{\sharp}-\m S^{\sharp}\m C_2\|_{\m F}^2\\
& \displaystyle + \frac{\beta}{2}\big(\|\m S^{\sharp}\|_{\m F}^2 - 2 \m T \m r (\m S^{\sharp})^{\m T}( \m J - \frac{\m Y_8}{\beta}) \big). \\
& \displaystyle \text{Adding $\|\m g(\m S) - \frac{\m Y_1}{\beta}\|_{\m F}^2$ and $\|\m J - \frac{\m Y_8}{\beta}\|_{\m F}^2$} \text{~~inside the first term and last term}\\
& \displaystyle \text{respectively to get the quadratic form. As these terms are constants when } \\
& \displaystyle \text{minimizing over $\m S^\sharp$ it will not affect the final solution.}\\
& \displaystyle \m S^{\sharp} = \underset{\m S^\sharp} {\text{argmin}} \frac{\beta}{2}\|\m S^{\sharp}-( \m g(\m S) - \frac{\m Y_1}{\beta})\|_{\m F}^2 + <\m Y_3, \m S^{\sharp}- \m S^{\sharp}\m C_2> + \frac{\beta}{2}\|\m S^{\sharp}-\m S^{\sharp}\m C_2\|_{\m F}^2 + \\
& \displaystyle \frac{\beta}{2}\|\m S^{\sharp} - (\m J-\frac{\m Y_8}{\beta})\|_{\m F}^2.
\end{aligned}
\label{eq:simple_form_Ssharp}
\end{equation}
The closed form solution for $\m S^\sharp$ can be derived by taking derivative of \eqref{eq:simple_form_Ssharp} w.r.t $\m S^\sharp$ and equating to zero.
\begin{equation}
\begin{aligned}
\m S^{\sharp}(\m 2\m I + (\m I-\m C_2)(\m I-\m C_{\m 2}^{\m T}) ) = \left(\m g(\m S) - \frac{\m Y_{1}}{\beta}\right) + (\m J - \frac{\m Y_8}{\beta}) - \frac{\m Y_3}{\beta}(\m I-\m C_{2}^{\m T}).
\end{aligned}
\end{equation}

\subsection{{Solution for $\m C_{1}$ }}
\begin{equation}
\begin{aligned}
& \displaystyle \m C_{1} = \underset{\m C_1} {\text{argmin}} < \m Y_2, \m S-\m S\m C_1> + \frac{\beta}{2}\|\m S-\m S\m C_1\|_{\m F}^2 + <\m Y_4, \m 1^{\m T} \m C_1-\m 1^{\m T}> +
\\
& \displaystyle \frac{\beta}{2}\|\m 1^{\m T}\m C_1-1^{\m T}\|_{\m F}^2 + <\m Y_6, \m C_1-\m E_1> + \frac{\beta}{2}\|\m C_1-\m E_1\|_{\m F}^2.\\
& \displaystyle = \underset{\m C_1} {\text{argmin}} \frac{\beta}{2}\|\m S\m C_1-(\m S + \frac{\m Y_2}{\beta})\|_{\m F}^2 + \frac{\beta}{2}\|\m 1^{\m T}\m C_1-(\m 1^{\m T} - \frac{\m Y_4}{\beta})\|_{\m F}^2 + \frac{\beta}{2}\|\m C_1-(\m E_1 - \frac{\m Y_6}{\beta})\|_{\m F}^2.
\end{aligned}
\end{equation}

The closed form solution for $\m C_{1}$ is solved as:
\begin{equation}
\begin{aligned}
(\m S^{\m T}\m S + \m 1 \m 1^{\m T} + \m I)\m C_{1} = \m S^{\m T}(\m S + \frac{\m Y_{2}}{\beta}) + \m 1(\m 1^{\m T}-\frac{\m Y_4}{\beta}) + (\m E_1 - \frac{\m Y_6}{\beta}).
\end{aligned}
\end{equation}

\begin{equation}
\m C_1 = \m C_1 - \mathrm{diag}(\m C_1),
\end{equation}

\subsection{{Solution for $\m C_{2}$ } }
\begin{equation}
\begin{aligned}
& \displaystyle \m C_{2} = \underset{\m C_2} {\text{argmin}} <\m Y_3, \m S^{\sharp}-\m S^{\sharp}\m C_2> + \frac{\beta}{2}\|\m S^{\sharp}-\m S^{\sharp}\m C_2\|_{\m F}^2 + <\m Y_5, \m 1^{\m T}\m C_2-\m 1^{\m T}> + \\
& \displaystyle + \frac{\beta}{2}\|\m 1^{\m T}\m C_2-\m 1^{\m T}\|_{\m F}^2 + <\m Y_7, \m C_2-\m E_2> + \frac{\beta}{2}\|\m C_2-\m E_2\|_{\m F}^2.
\\
& \displaystyle = \underset{\m C_2} {\text{argmin}}  \frac{\beta}{2}\|\m S^{\sharp}\m C_2-(\m S^{\sharp} + \frac{\m Y_3}{\beta})\|_{\m F}^2 + \frac{\beta}{2}\|\m 1^{\m T}\m C_2-(\m 1^{\m T} - \frac{\m Y_5}{\beta})\|_{\m F}^2 + \frac{\beta}{2}\|\m C_2-(\m E_2 - \frac{\m Y_7}{\beta})\|_{\m F}^2.
\end{aligned}
\end{equation}

The closed form solution for $\m C_{2}$ is derived as:
\begin{equation}
\begin{aligned}
\left(({\m S^{\sharp}})^{\m T}\m S^{\sharp} + \m 1 \m 1^{\m T} + \m I\right)\m C_{2} = ({\m S^{\sharp}})^{T}\m (S^{\sharp} + \frac{\m Y_{3}}{\beta}) + \m 1(1^{\m T}-\frac{\m Y_5}{\beta}) + (\m E_2 - \frac{\m Y_7}{\beta}).
\end{aligned}
\end{equation}

\begin{equation}
\m C_2 = \m C_2 - \mathrm{diag}(\m C_2),
\end{equation}

\subsection{Solution for $\m E_{1}$}
\begin{equation}
\begin{aligned}
& \displaystyle \m E_1 = \underset{\m E_1}{\text{argmin}} \lambda_{1}\|\m E_1 \|_1 + \gamma_{1}\|\m E_1\|_{\m F}^2 + <\m Y_6, \m C_1-\m E_1> + \frac{\beta}{2}\|\m C_1 - \m E_1\|_{\m F}^2. \\
& \displaystyle  = \underset{\m E_1}{\text{argmin}} \lambda_{1}\|\m E_1\|_1 + \gamma_{1}\|\m E\|_{\m F}^2 + \frac{\beta}{2} \|\m E_1 -(\m C_1 + \frac{\m Y_6}{\beta})\|_{\m F}^2. \\
& \displaystyle  = \underset{\m E_1}{\text{argmin}} \lambda_{1}\|\m E_1\|_1 + \gamma_{1}\|\m E_1\|_{\m F}^2 + \frac{\beta}{2} \|\m E_1\|_{\m F}^2 - \beta<\m E_1, (\m C_1 + \frac{\m Y_6}{\beta})>\\
& \displaystyle = \underset{\m E_1}{\text{argmin}} \lambda_{1}\|\m E_1\|_1 + (\gamma_1 + \frac{\beta}{2})(\|\m E_1\|_{\m F}^2 + \frac{2\beta}{2\gamma_1 + \beta} <\m E_1, \m C_1 + \frac{\m Y_6}{\beta}>).\\
& \displaystyle = \underset{\m E_1}{\text{argmin}} \lambda_{1}\|\m E_1\|_1 + (\gamma_1 + \frac{\beta}{2})\|\m E_1 - \frac{\beta}{2\gamma_1 + \beta}(\m C_1 + \frac{\m Y_6}{\beta})\|_{\m F}^2.\\
\end{aligned}
\end{equation}

The closed form solution for $\m E_1$ is reached as:
\begin{equation}
\m E_{1} = \mathcal{S}_{\frac{\lambda_1}{\gamma_1 + \frac{\beta}{2}}}(\frac{\beta}{2\gamma_1 + \beta}(\m C_1 + \frac{\m Y_6}{\beta}))
\end{equation}

\subsection{Solution for $\m E_{2}$ }
The derivation for the solution of $\m E_2$ is similar to the solution of $\m E_1$.
\begin{equation}
\begin{aligned}
& \displaystyle \m E_2 = \underset{\m E_2}{\text{argmin}} \lambda_{3}\|\m E_2 \|_1 + \gamma_{3}\|\m E_2\|_F^2 + <\m Y_7, \m C_2-\m E_2> + \frac{\beta}{2}\|\m C_2 - \m E_2\|_{\m F}^2 \\
& \displaystyle = \underset{\m E_2}{\text{argmin}} \lambda_{3}\|\m E_2\|_1 + (\gamma_3 + \frac{\beta}{2})\|\m E_2 - \frac{\beta}{2\gamma_3 + \beta}(\m C_2 + \frac{\m Y_7}{\beta})\|_{\m F}^2.\\
\end{aligned}
\end{equation}
The closed form solution for $\m E_2$ is reached as:
\begin{equation}
\m E_{2} = \mathcal{S}_{\frac{\lambda_3}{\gamma_3 + \frac{\beta}{2}}}(\frac{\beta}{2\gamma_3 + \beta}(\m C_2 + \frac{\m Y_7}{\beta})).
\end{equation}

\section{Tables for each comparison}\label{ap:appendixB}
\begin{table}
\centering
\caption{Table corresponding to Figure \ref{fig:synthetic_comparison}}
\label{tab:Btable1}
\begin{tabular}{|c|c|c|c|c|c|}\hline
Datasets      & BMM   & PND            & Zhu et al. & Kumar et al. & Ours           \\\hline
Dance+Yoga    & 0.045 & 0.078          & 0.052      & 0.046        & \textbf{0.043} \\\hline
Drink+Walking & 0.074 & \textbf{0.060} & 0.083      & 0.073        & 0.071          \\\hline
Shark+Stretch & 0.024 & \textbf{0.015} & 0.067      & 0.025        & 0.019          \\\hline
Walking+Yoga  & 0.070 & 0.072          & 0.087      & 0.070        & \textbf{0.066} \\\hline
Face+Pickup   & 0.032 & \textbf{0.012} & 0.018      & 0.025        & 0.022          \\\hline
Face+Yoga     & 0.017 & \textbf{0.010} & 0.028      & 0.019        & 0.017          \\ \hline
Shark+Yoga    & 0.035 & \textbf{0.018} & 0.094      & 0.037        & 0.033          \\\hline
Stretch+Yoga  & 0.039 & 0.109          & 0.045      & 0.039        & \textbf{0.036}\\
\hline
\end{tabular}
\end{table}

\begin{table}
\centering
\caption{Table corresponding to Figure \ref{fig:reconstructionErrorUMPM}}
\label{tab:Btable2}
\begin{tabular}{|c|c|c|c|c|c|}
\hline
Datasets      & BMM    & PND    & Zhu et al.      & Kumar et al. & Ours            \\ \hline
p2\_free\_2   & 0.1973 & 0.1544 & \textbf{0.1142} & 0.1992       & 0.1171          \\ \hline
p2\_grab\_2   & 0.2018 & 0.1570 & 0.0960          & 0.2080       & \textbf{0.0822} \\ \hline
p3\_ball\_1   & 0.1356 & 0.1477 & 0.0832          & 0.1348       & \textbf{0.0810} \\ \hline
p4\_meet\_12  & 0.0802 & 0.0862 & 0.0972          & 0.0821       & \textbf{0.0815} \\ \hline
p4\_table\_12 & 0.2313 & 0.1588 & 0.1322          & 0.2313       & \textbf{0.0994} \\ \hline
\end{tabular}
\end{table}

\begin{table}[H]
\centering
\caption{Table corresponding to Figure \ref{fig:dense_3d_synthetic} }
\label{tab:Btable3}
\begin{tabular}{|c|c|c|c|c|c|}
\hline
Datasets        & BMM   & PND   & Zhu et al. & Kumar et al.   & Ours           \\ \hline
Face Sequence 1 & 0.078 & 0.077 & 0.082      & 0.075          & \textbf{0.073} \\ \hline
Face Sequence 2 & 0.059 & 0.062 & 0.063      & \textbf{0.050} & 0.052          \\ \hline
Face Sequence 3 & 0.042 & 0.051 & 0.057      & \textbf{0.038} & 0.039          \\ \hline
Face Sequence 4 & 0.049 & 0.041 & 0.056      & 0.044          & \textbf{0.040} \\ \hline
\end{tabular}
\end{table}
\end{document}